\documentclass{article} % For LaTeX2e
\usepackage{iclr2027_conference,times}

\usepackage[utf8]{inputenc} % allow utf-8 input
\usepackage[T1]{fontenc}    % use 8-bit T1 fonts
\usepackage{hyperref}       % hyperlinks
\usepackage[capitalize]{cleveref}
\usepackage{url}            % simple URL typesetting
\usepackage{booktabs}       % professional-quality tables
\usepackage{amsfonts}       % blackboard math symbols
\usepackage{nicefrac}       % compact symbols for 1/2, etc.
\usepackage{microtype}      % microtypography
\usepackage[dvipsnames]{xcolor}         % colors
\usepackage{subcaption}
\usepackage{enumitem}
\usepackage{tikz}
\usepackage{tikz-cd}
\usetikzlibrary{cd,arrows,calc,shapes,positioning}
\usepackage{wrapfig}
\usepackage{graphicx}
\usepackage{bbm}

\newcommand{\xhdr}[1]{\textbf{#1}\:}
\usepackage{amsthm}

\newtheorem{theorem}{Theorem}
\newtheorem{proposition}[theorem]{Proposition}
\theoremstyle{definition}

\usepackage{thmtools,thm-restate}

\usepackage{acronym}
\acrodef{IV}{Instrumental Variable}
\acrodef{OU}{Ornstein--Uhlenbeck}
\acrodef{SDE}{Stochastic Differential Equation}
\acrodef{KBE}{Kolmogorov--Backward Equation}
\acrodef{MLE}{Maximum-Likelihood Estimation}

\tikzstyle{obs} = [circle,fill=white,draw=black,inner sep=1pt,minimum size=25pt,node distance=0.75cm,thick]%font=\fontsize{10}{10}\selectfont,
\tikzstyle{latent} = [obs,dotted]
\tikzstyle{fixed} = [obs,rectangle,minimum size=22pt]
\tikzstyle{small} = [inner sep=0pt,minimum size=20pt]%font=\fontsize{10}{10}\selectfont,

\newcommand{\edge}[3][]{ %
  \foreach \x in {#2} { %
    \foreach \y in {#3} { %
      \path (\x) edge [->, >={triangle 45}, thick, #1] (\y) ;%
    } ;
  } ;
}

\iclrfinalcopy

\title{Instrumental Variables in Continuous-Time Stochastic Processes}

\author{Lars Pennig \\
Technical University of Munich \\
Helmholtz Munich \\
Munich Center for Machine Learning \\
\texttt{lars.pennig@tum.de} \\
\And
Søren Wengel Mogensen \\
Center for Statistics, Department of Finance \\
Copenhagen Business School \\
\texttt{swm.fi@cbs.dk} \\
\AND
Niki Kilbertus \\
Technical University of Munich \\
Helmholtz Munich \\
Munich Center for Machine Learning \\
\texttt{niki.kilbertus@tum.de} \\
}

\begin{document}

\maketitle
\lhead{Preprint}

\begin{abstract}
Causal questions often arise in systems that evolve in continuous time. While instrumental variable (IV) methods provide a principled approach to identify causal effects in static or discrete-time settings, even in the presence of unobserved confounding, they do not yet extend to continuous-time dynamics. In this work, we develop an IV framework for stationary continuous-time stochastic processes parameterized by stochastic differential equations (SDE), where the instrument itself takes the form of a stochastic process. We encode the structural sparsity and causal effects directly at the infinitesimal level of the drift, and contrast these assumptions with the IV assumptions on the discretized version of the same process. Based on these structural assumptions, we derive moment restrictions using path-valued instruments. We show how these restrictions simplify for Ornstein-Uhlenbeck processes and present an identification condition together with a corresponding consistent estimation method. Finally, we extend our estimation framework to nonlinear causal effects by expressing the path-based moment restrictions leveraging the signature kernel. Empirically, our method recovers causal effects in confounded continuous-time systems where maximum-likelihood-based estimation and discrete-time IV methods are systematically biased.
\end{abstract}

\section{Introduction}
Causal effect estimation under unobserved confounding is a central challenge in many scientific domains. In the presence of unobserved confounders, standard estimation techniques fail to distinguish between confounding and the true effect. This distinction is generally impossible from observational data without additional assumptions. The \ac{IV} framework forms one of the most prominent settings in which principled strategies for cause-effect identification and estimation exist under certain assumptions. \acp{IV} are exogenous variables that affect the covariates, are independent of the hidden confounders, and only affect the outcome through the covariates. \ac{IV} methods have been thoroughly studied and are an established workhorse in the toolbox of econometricians~\citep{sargan_iv,Hansen_GMM, iv_econ} and beyond. They also attracted renewed interest in the context of flexible machine learning techniques~\citep{bennett2020deepgeneralizedmethodmoments, hartford_deep_iv, ailer2025targetedsequentialindirectexperiment}.

While a range of methods exists for the static IV setting, many real-world systems in which questions of causation arise are inherently dynamic, including applications in earth system sciences~\citep{Runge2019, Stips2016}, medical settings~\citep{hernan2001,Lok_2008} and macroeconomic analyses \citep{fernandez2006estimating_nber, sims2002solving}. 
As an example, consider the objective of identifying the glucose-insulin sensitivity of a type 2 diabetes patient, which describes the main mechanism impaired by the disease. Insulin and glucose concentration dynamics may be confounded by unobserved digestive processes, rendering standard effect estimation techniques biased. Using the IV structure of the problem, insulin sensitivity may still be identified using the external insulin administration process as an instrument. We will get back to this example in our empirical validation.
%As an example, consider the objective of identifying the effect of a time-varying drug concentration $X$ on a disease marker $Y$, such as a tumor size or viral load. Both processes are confounded by the patient's latent immune response dynamics, which affect drug absorption as well as disease progression directly. The true effect of the drug may still be identified by using the drug administration process as an instrument.
Recent work on \ac{IV} methods for parameter estimation of discrete-time processes~\cite{mogensen2023, thams2024} is unsuitable for continuous-time systems. First, observational data is often available only on irregular sampling-intervals, which discrete-time methods fail to handle. Moreover, if the underlying real world system evolves in continuous-time, the structural IV sparsity needs to be modeled on an \emph{infinitesimal scale} rather than over finite time intervals. That is, discrete-time models imply structural IV assumptions that do not preserve the actual local IV structure of a continuous-time process, resulting in severe model-misspecification.

Furthermore, our continuous-time formulation has strictly more ``identification power'' than in the static case. Concretely, in static IV settings identification is fundamentally limited by the variation and number of available instruments. For example, in the linear case, identification requires as many instruments as treatments. The relevant underspecified case, where only few instruments are available compared to treatments, has been acknowledged as an important setting~\citep{ailer2023sequentialunderspecifiedinstrumentselection}, but received little attention. In contrast, in our new continuous-time setting, we show that the temporal structure of the process contains richer identifying information, allowing for identification of multi-dimensional causal effects from a single instrument process. 

We summarize our \textbf{contributions} as follows:
\begin{itemize}[leftmargin=*,itemsep=-0.5mm,topsep=-1mm]
    \item We are first to formulate the IV setting in continuous-time stochastic processes parametrized by SDEs, where the IV structure is defined on the infinitesimal level of the process.%While highly relevant, this setting has not yet been defined or discussed in the literature.
    \item We derive the moment conditions implied by the IV structure in stationary SDE systems as orthogonality between forecasting errors and functionals of the instrumental process path.
    \item We derive a controllability condition that \emph{identifies} the target parameters in a confounded \ac{OU} process, along with a \emph{consistent} estimator.
    \item We propose an extension for nonlinear objective functions via a kernel-based estimation scheme.
    \item We demonstrate empirically that our method outperforms possible alternative approaches.
\end{itemize}

\section{Static and Discrete-Time Instrumental Variables}

\xhdr{Static IV.} 
In the static setting, the standard IV setup aims to identify the effect of covariates $X$ on outcomes $Y$ under confounding of an unobserved variable $U$ by making use of the instrument variable $Z$. In general, IV is defined by three assumptions on the underlying data-generating mechanism, \textit{relevance, exclusion and exogeneity}~\citep{hernan2020causal, Pearl_causality}. Relevance requires that $X$ is not independent of $Z$ ($X \not\!\perp\!\!\!\perp Z$), the exclusion restriction requires that $Z$ only affects $Y$ through $X$ ($Z \perp\!\!\!\perp Y \mid X, U$) and exogeneity requires that $U$ is independent of $Z$ ($U\!\perp\!\!\!\perp Z$). The standard ``IV graph'' encoding these conditions is illustrated in~\cref{fig:standard_iv_scm}.

Instrumental variable methods and generalized-method-of-moments (GMM) frameworks use the exogeneity of instrument variables together with setting-specific assumptions to formulate effect-identifying moment conditions, and construct associated estimators~\citep{sargan_iv,Hansen_GMM,wright1928tariff}. Albeit the IV setting being restrictive as valid instruments are often hard to find, these methods continue to uncover relevant causal effects, such as the medical cost of air pollution, effects of medical aid programs, among other~\citep{IV_example_1,IV_example_2,IV_example_3,IV_example_4}.
In recent years, deep learning methods have been applied to both IV- and GMM-setups
\citep{bennett2020deepgeneralizedmethodmoments, saengkyongam2022exploitingindependentinstrumentsidentification, hartford_deep_iv} and in a different line of work, kernel-based methods are used to approximate causal effects~\citep{ailer2025targetedsequentialindirectexperiment, bennett2023minimaxinstrumentalvariableregression, zhang2023instrumentalvariableregressionkernel}

\xhdr{IV methods for discrete processes.} Recent papers have described instrumental variable techniques in stationary multivariate vector autoregressive processes in discrete time \citep{mogensen2023,thams2024}. \cite{mogensen2023} proposes using a time-integrated covariance matrix to discover \textit{normalized parameters}, which can be given a causal interpretation. This also extends to linear Hawkes processes, which describe continuous-time event processes \citep{mogensen2023}. \cite{thams2024} investigate discrete-time instrumental processes by studying graphical properties of the time-unrolled causal graph of the process. They clarify that the previously described static IV conditions are not trivially fulfilled in such a graph, as past instruments violate exclusion by causally affecting both current instrument states and outcomes. They propose two approaches for estimating causal effects in this setting: \textit{conditional IV}, which retrieves the standard IV setting by conditioning on subsets of the process history, and \textit{nuisance IV}, which adds past outcomes to the covariates.

\xhdr{Causal interpretation of SDEs.} A growing amount of papers seeks causal interpretations of SDEs, e.g., \cite{HansenSokhol_2014, manten2025signaturekernelconditionalindependence, lorch_sdes,schneider2026conservative}, by defining interventions in SDE systems and performing causal discovery of the local independence graph in SDE systems. We extend this line of work by contributing IV-type estimation for continuous-time processes defined by SDEs.

\xhdr{Instrumental variables in system identification.} Adjacent to our setting, IV methods can also be found in the systems engineering and control theory literature, mostly in the context of discrete-time stochastic processes, starting from early papers such as
\cite{joseph1961plant,wong1967identification}. This line of work often assumes an explicit \emph{control input} which may serve as an instrumental process, alternatively lagged variables are used as instruments \citep{friedlander1985instrumental,gilson2005instrumental}. The goal is most often the identification of a state-space model, instead of the identification of a set of causal parameters.

\section{Preliminaries}\label{sec:preliminaries}

\xhdr{Setup.} We first introduce a general, continuous-time, stochastic data-generating mechanism in which we can encode the structural IV assumptions. Specifically, the data is generated by a stochastic process with states $\mathbf{S}_t = (Y_t, X_t, Z_t, U_t) \in \mathbb{R}^{d_y + d_x + d_z + d_u}$ indexed by time $t$ and driven by a Wiener process $\mathbf{W}_t = (W_t^y,W_t^x,W_t^z,W_t^{u})$ with jointly independent components. Process $Z$ acts as an instrument process, $X$ represents the evolving covariates, $Y$ the outcome process, and $U$ the unobserved confounding process.
All causal relationships between the variables are encoded in the structure of the drift, which allows for a direct interpretation of the causal effects, in contrast to diffusion coupling, and avoids additional identifiability challenges. Thus, we model the dynamics via the following system of SDEs 
\begin{equation}
d\mathbf{S}_t =
\underbrace{
\begin{pmatrix} 
{\color{MidnightBlue}f_y(Y_t,X_t)} + U_t \\[1mm]
f_x(Y_t,X_t,Z_t,U_t) \\[1mm]
f_z(Z_t) \\[1mm]
f_u(U_t)
\end{pmatrix}}_{\mathbf{f}(\mathbf{S}_t)}
\,dt
+
\underbrace{
\begin{pmatrix} 
\sigma_y(Y_t) & 0 & 0 & 0 \\ 
0 & \sigma_x(X_t) & 0 & 0 \\ 
0 & 0 & \sigma_z(Z_t) & 0 \\ 
0 & 0 & 0 & \sigma_u(U_t)
\end{pmatrix}}_{\boldsymbol{\sigma}(\mathbf{S}_t)}
\begin{pmatrix} dW^y_t \\ dW^x_t \\ dW^z_t \\ dW^u_t \end{pmatrix}
\label{eq:drift_confounded_sde}
\end{equation}
with $\mathbf{S}_0 \sim \pi$, where $\pi$ is the unique invariant measure of the process. We assume drift $\mathbf{f}$ and diffusion $\boldsymbol{\sigma}$ to be globally Lipschitz and satisfy linear growth conditions, ensuring existence and pathwise uniqueness of a strong solution to \cref{eq:drift_confounded_sde}~\citep{pavliotis2014stochastic}, as well as finite first moments $ \mathbb{E}_{\pi}\!\left[\|\mathbf{S}\|\right] < \infty$. The resulting joint process is assumed to be strongly stationary. We assume without loss of generality that the process is zero-mean. All components of $\mathbf{S}_t$ aside from $U_t$ are observed. In \cref{app:setup}, we discuss the assumptions and alternative confounding types based on non-diagonal diffusion matrices, for which our proposed methods remain applicable.

\begin{wrapfigure}{r}{0.55\columnwidth}
  \vspace{-12pt}
  \centering
  \resizebox{0.5\columnwidth}{!}{% Full composite figure for the paper, all in TikZ.
%   Left  : confounded continuous-time process trajectories
%   Right : local independence graph of structural IV assumptions
%   Bottom: converging arrows pointing to the identification objective
%
% Compile with: pdflatex iv_figure.tex
% Trajectory coordinates live in trajectory_data.tex (generated by Python).

% \documentclass[border=8pt]{standalone}
% \usepackage{amsmath, amssymb}
% \usepackage{xcolor}
% \usepackage{tikz}
% \usetikzlibrary{arrows.meta, positioning, calc, bending}

% \definecolor{accent}{RGB}{90,110,140}    % slate-blue
% \definecolor{cY}{HTML}{3a72b8}
% \definecolor{cU}{HTML}{7a7a7a}
% \definecolor{cX}{HTML}{e07b3c}
% \definecolor{cZ}{HTML}{3a8f4d}

\input{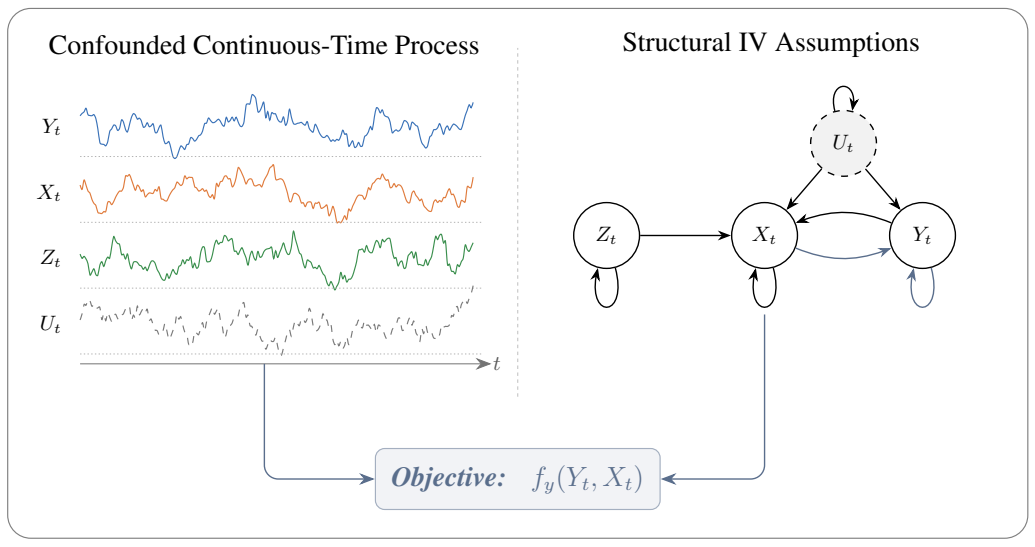}              % defines \trajY, \trajU, \trajX, \trajZ

\begin{tikzpicture}[
    >={Stealth[length=2.4mm]},
    every node/.style={font=\small},
    obs/.style       = {circle, draw, semithick, minimum size=10mm,
                        inner sep=0pt, font=\small},
    latent/.style    = {circle, draw, semithick, dashed, minimum size=10mm,
                        inner sep=0pt, font=\small, fill=gray!10},
    causal/.style    = {-{Stealth[length=2mm]}, semithick},
    feedback/.style  = {-{Stealth[length=2mm]}, semithick, bend right=22},
    target/.style    = {-{Stealth[length=2mm]}, semithick, color=accent},
    targetfb/.style  = {-{Stealth[length=2mm]}, semithick, color=accent, bend right=22},
  ]
 
% =========================================================================
% Outer rounded rectangle + panel divider
% =========================================================================
\draw[rounded corners=10pt, line width=0.5pt, color=black!50]
  (-0.5, -2.20) rectangle (15.0, 5.85);
 
% Subtle dashed separator between the two panels
\draw[line width=0.4pt, color=black!25, dash pattern=on 1.5pt off 1.5pt]
  (7.25, -0.05) -- (7.25, 5.25);
 
% =========================================================================
% LEFT PANEL — trajectories
% =========================================================================
\node[anchor=center, font=\large]
  at (3.4, 5.30) {Confounded Continuous-Time Process};
 
% Each trajectory: label (anchor=east), curve, and a faint baseline.
\newcommand{\trajbaseline}[5]{%
  \node[anchor=east] at (0.45, #1) {#2};%
  \begin{scope}[xshift=0.6cm, yshift=#1 cm]
    \draw[#5, line width=0.5pt, color=#3] plot[smooth] coordinates {#4};
  \end{scope}
  \draw[line width=0.35pt, color=black!30, dash pattern=on 0.6pt off 1pt]
    (0.6, #1-0.45) -- (6.7, #1-0.45);
}
 
\trajbaseline{4.05}{$Y_t$}{cY}{\trajY}{solid}
\trajbaseline{3.05}{$X_t$}{cX}{\trajX}{solid}
\trajbaseline{2.05}{$Z_t$}{cZ}{\trajZ}{solid}
\trajbaseline{1.05}{$U_t$}{cU}{\trajU}{dashed}
 
% Time axis (real arrow, with label)
\draw[->, line width=0.4pt, color=black!55] (0.6, 0.45) -- (6.85, 0.45);
\node[anchor=west, color=black!65] at (6.75, 0.45) {$t$};
 
% =========================================================================
% RIGHT PANEL — local independence graph
% =========================================================================
\node[anchor=center, font=\large]
  at (11.1, 5.30) {Structural IV Assumptions};
 
% Wider horizontal spread (2.4 instead of 1.6) and centred vertically
% so the panel actually fills its half of the figure.
\begin{scope}[shift={(8.6, 2.40)}]
  \node[obs]    (Z) at (0,   0)   {$Z_t$};
  \node[obs]    (X) at (2.4, 0)   {$X_t$};
  \node[obs]    (Y) at (4.8, 0)   {$Y_t$};
  \node[latent] (U) at (3.6, 1.40){$U_t$};
 
  % Self-loops (drift). Smaller loop on U so it doesn't reach the title.
  \draw[causal] (Z) edge[loop below, looseness=8] (Z);
  \draw[causal] (X) edge[loop below, looseness=8] (X);
  \draw[target] (Y) edge[loop below, looseness=8] (Y);
  \draw[causal] (U) edge[loop above, looseness=5] (U);
 
  % Cross-effects
  \draw[causal]   (Z) -- (X);
  \draw[feedback] (Y) to (X);
  \draw[targetfb] (X) to (Y);
  \draw[causal]   (U) -- (X);
  \draw[causal]   (U) -- (Y);
\end{scope}
 
% =========================================================================
% BOTTOM — Objective callout with converging arrows
% =========================================================================
\node[font=\large, color=accent,
      draw=accent!55, line width=0.5pt,
      rounded corners=4pt, fill=accent!8,
      inner sep=7pt] (obj)
  at (7.25, -1.30) {\textbf{\textit{Objective:}}\;\; $f_y(Y_t, X_t)$};
 
% Converging arrows: anchored to a clear point in each panel,
% routed with rounded corners into the callout box.
\draw[->, semithick, color=accent, rounded corners=6pt]
  (3.4, 0.45) -- (3.4, -1.30) -- (obj.west);
\draw[->, semithick, color=accent, rounded corners=6pt]
  (11.0, 1.20) -- (11.0, -1.30) -- (obj.east);
 
\end{tikzpicture}}
  \caption{Observed trajectories of processes $(Y, X, Z)$,
    confounded by the hidden process $U$, together with the
    assumed sparse drift graph, are used to identify
    the objective function {\color{MidnightBlue}$f_y(Y_t,X_t)$}.}
  \label{fig:iv-setup}
  \vspace{-9pt}
\end{wrapfigure}
% Generalize IV assumptions to continuous time 
The structural assumptions implied by~\cref{eq:drift_confounded_sde} directly parallel the static IV assumptions and are illustrated in the sparsity graph of the process, shown in~\Cref{fig:iv-setup}. First, we assume \textit{relevance}, meaning that the drift of process $X$ depends non-trivially on process $Z$. Second, the \textit{exclusion restriction} requires that the instrument process only influences the outcome process $Y$ through the covariate process $X$, thus there is no direct effect of process $Z$ on $Y$ ($Z$ is not an argument of $f_y$). 
Third, \emph{exogeneity} in continuous-time implies that the instrument process $Z_t$ is independent of the Brownian motion increments $dW_t^y$ as well as of the confounding process $U_t$. The latter is described by the following conditional moment restriction
\begin{equation}
\mathbb{E}\left[U_t \mid \mathcal{F}^z_s\right] = 0, \quad \text{for } s \leq t,
\label{eq:orthogonal_instruments}
\end{equation}

where $\mathcal{F}^z_s := \sigma(Z_{[0,s]})$ denotes the filtration of the instrument process up to time $s$. Analogously to most static IV settings, we assume that $U_t$ enters the drift of the outcome process $Y$ \textit{additively}~\citep{hernan2020causal}. While restrictive, without additivity one typically needs similarly strong alternative assumptions to achieve identifiability~\citep{balke,Imbens}.

\xhdr{Data.} We observe a finite trajectory of the marginal process and record tuples $(Y_{t_i}, X_{t_i}, Z_{t_i}, t_i)$ at $n$ observation times $0\le t_1<t_2<\ldots < t_n \le T_{max}$. The states are all measured simultaneously, however, the spacing between timepoints may be irregular. The confounding process $U$ is latent and never observed. In principle, a single long trajectory suffices for identification and estimation. For practical purposes, in our empirical evaluation we sample multiple shorter trajectories for efficiency.

\xhdr{Objective.} The main objective of this paper is the identification of the drift component $f_{y}(Y_t,X_t)$ describing the causal effect of covariate process $X$ on outcome process $Y$ and the self-dynamics of $Y$. Although the causal effect of process $X$ on process $Y$ is the primary effect of interest, it is inherently entangled with the self-dynamics of $Y$ and both are therefore jointly identified and estimated.

% Explain why we need IV methods - show that mle for sdes fails
\xhdr{Bias in MLE.} In the given setting, maximum likelihood estimation of $f_{y}$ is biased due to unobserved confounding. This becomes evident by analyzing the transition densities of the partially observed process. Naive \ac{MLE} maximizes $\mathbb{E}\left[ \log p_\theta(Y_{t+\Delta t} \mid X_t, Y_t)\right]$ under some approximation of the transition density~\citep{pavliotis2014stochastic}. This transition density depends on the unobserved $U_t$ via
% \begin{equation}
$
    p(Y_{t+\Delta t} \mid X_t, Y_t) = \int p(Y_{t+\Delta t} \mid X_t, Y_t, U_t)\, p(U_t \mid X_t, Y_t)\, dU_t.
$
% \end{equation}
Since $U_t$ confounds both $X_t$ and $Y_t$, a likelihood model omitting $U_t$ generally attributes part of the confounding mechanism to the causal effect $f_{y}$.
This motivates the use of IV methods that exploit the structure to identify the drift function.

% What kind of additional requirements / challenges come with IV in continuous time vs the static setting and the discrete-time setting?
\xhdr{Relation to other IV settings.}
% In contrast to the static and discrete-time setting, the setup described in \cref{eq:drift_confounded_sde} implies additional challenges for IV methods.
Unlike in the static case, the outcome $Y$ depends on its own past and may interact cyclically with process $X$ leading to substantially different structure of the objective function $f_y$. 
Compared to the discrete-time dynamic setting, the local independencies implied by the continuous-time setup in \cref{eq:drift_confounded_sde} differ fundamentally from structural assumptions in discrete-time processes described in~\cite{mogensen2023,thams2024}. In continuous time, the structural sparsity exists only on the infinitesimal level and does not imply an identical sparsity pattern in finite-step transition maps of the corresponding discretely observed process~\citep{Mogensen2018CausalLF}.
For example, an equidistantly sampled \ac{OU} process admits an equivalent representation as a $\mathrm{VAR}(1)$ process. However, when imposing an IV structure on the drift of the OU process, the exclusion restriction is generally not preserved in the corresponding $\mathrm{VAR}(1)$ transition matrix. The key is that sparsity in an infinitesimal generator matrix $A$ does not transfer to the exponential map $e^{A \Delta t}$ for a finite interval (see~\cref{app:discrete_vs_conti_independencies} for detail), e.g., for $\Delta t = 1$:
\begin{center}
\vspace{-3mm}
\resizebox{!}{1cm}{\tikzset{
  heatmatrix/.style={
    matrix of math nodes,
    left delimiter=(,
    right delimiter=),
    nodes in empty cells,
    nodes={
      minimum width=1.00cm,
      minimum height=0.54cm,
      inner sep=0pt,
      outer sep=0pt,
      anchor=center,
      text height=1.6ex,
      text depth=.4ex
    },
    column sep=1pt,
    row sep=1pt
  },
  hmlabel/.style={text=gray, font=\scriptsize}
}

\newcommand{\AddMatrixLabels}[1]{%
  \node[hmlabel, anchor=south] at ([yshift=5pt]#1-1-1.north) {$Y_t$};
  \node[hmlabel, anchor=south] at ([yshift=5pt]#1-1-2.north) {$X_t$};
  \node[hmlabel, anchor=south] at ([yshift=5pt]#1-1-3.north) {$Z_t$};
  \node[hmlabel, anchor=south] at ([yshift=5pt]#1-1-4.north) {$U_t$};

  \node[hmlabel, anchor=west] at ($ (#1.east |- #1-1-4.center) + (7pt,0) $) {$Y_t$};
  \node[hmlabel, anchor=west] at ($ (#1.east |- #1-2-4.center) + (7pt,0) $) {$X_t$};
  \node[hmlabel, anchor=west] at ($ (#1.east |- #1-3-4.center) + (7pt,0) $) {$Z_t$};
  \node[hmlabel, anchor=west] at ($ (#1.east |- #1-4-4.center) + (7pt,0) $) {$U_t$};
}

\newcommand{\Zero}{|[fill=black!10]|}

\newcommand{\NegA}{|[fill=BrickRed!34]|}
\newcommand{\NegB}{|[fill=BrickRed!27]|}
\newcommand{\NegC}{|[fill=BrickRed!20]|}
\newcommand{\NegD}{|[fill=BrickRed!13]|}
\newcommand{\NegE}{|[fill=BrickRed!8]|}

\newcommand{\PosA}{|[fill=RoyalBlue!34]|}
\newcommand{\PosB}{|[fill=RoyalBlue!27]|}
\newcommand{\PosC}{|[fill=RoyalBlue!20]|}
\newcommand{\PosD}{|[fill=RoyalBlue!13]|}
\newcommand{\PosE}{|[fill=RoyalBlue!8]|}

  \begin{tikzpicture}[baseline=(current bounding box.center)]
     \node[obs] (X1) {$Z_t$};
     \node[obs, right =of X1] (X2) {$X_t$};
     \node[obs, right =of X2] (X3) {$Y_t$};
     \node[latent, above = of $(X2)!0.5!(X3)$] (X4) {$U_t$};
     \edge{X1}{X2}
     \edge[out=15, in=165]{X2}{X3}
     \edge[out=195, in=-15]{X3}{X2}
     \edge{X4}{X2,X3}
     \draw[->, >={triangle 45}, thick] (X1) edge[loop below, looseness=8] (X1);
     \draw[->, >={triangle 45}, thick] (X2) edge[loop below, looseness=8] (X2);
     \draw[->, >={triangle 45}, thick] (X3) edge[loop below, looseness=8] (X3);
     \draw[->, >={triangle 45}, thick] (X4) edge[loop above, looseness=8] (X4);
  \end{tikzpicture}
  \hspace{1cm}

$A =$
\begin{tikzpicture}[baseline=(A.center),scale=0.8]
\matrix (A) [heatmatrix] {
\NegA -1.4  & \NegC -1   & \Zero 0     & \NegD -0.5 \\
\NegE -0.25 & \NegB -1.2 & \NegC -1    & \NegD -0.5 \\
\Zero 0     & \Zero 0    & \NegC -1    & \Zero 0 \\
\Zero 0     & \Zero 0    & \Zero 0     & \NegC -1 \\
};
\AddMatrixLabels{A}
\end{tikzpicture}
\quad \text{but} \quad
$e^{A\Delta t} \approx$
\begin{tikzpicture}[baseline=(B.center)]
\matrix (B) [heatmatrix] {
\PosC 0.280   & \NegC -0.285 & \PosD {\color{BrickRed}\textbf{0.154}} & \NegE -0.081 \\
\NegE -0.071  & \PosB 0.337  & \NegB -0.346 & \NegD -0.154 \\
\Zero 0       & \Zero 0      & \PosA 0.368  & \Zero 0 \\
\Zero 0       & \Zero 0      & \Zero 0      & \PosA 0.368 \\
};
\AddMatrixLabels{B}
\end{tikzpicture}

  \hspace{1cm}
  \begin{tikzpicture}[baseline=(current bounding box.center)]
     \node[obs] (X1) {$Z_t$};
     \node[obs, right =of X1] (X2) {$X_t$};
     \node[obs, right =of X2] (X3) {$Y_t$};
     \node[latent, above = of $(X2)!0.5!(X3)$] (X4) {$U_t$};
     \edge{X1}{X2}
     \edge[out=15, in=165]{X2}{X3}
     \edge[out=195, in=-15]{X3}{X2}
     \edge{X4}{X2,X3}
     \draw[->, >={triangle 45}, thick] (X1) edge[loop below, looseness=8] (X1);
     \draw[->, >={triangle 45}, thick] (X2) edge[loop below, looseness=8] (X2);
     \draw[->, >={triangle 45}, thick] (X3) edge[loop below, looseness=8] (X3);
     \draw[->, >={triangle 45}, thick] (X4) edge[loop above, looseness=8] (X4);
     \draw[->, >={triangle 45}, thick, BrickRed] (X1) edge[loop above, out=90, looseness=2] (X3);
  \end{tikzpicture}}
\vspace{-3mm}
\end{center}
Hence, when causal mechanisms operate in continuous-time with IV sparsity imposed on the infinitesimal level, \emph{discrete-time IV models are misspecified by construction}. 

\section{Continuous-Time IV Moment Conditions}
% \xhdr{Lagged conditional moment restrictions.}
We now derive the moment conditions implied by the IV structure using the independence between instrument and confounding processes described in~\cref{eq:orthogonal_instruments}.

\begin{proposition}[Continuous-time IV Moment Restriction]
    \label{prop:moment_conditions}
    %\label[proposition]{prop:moment_conditions}
Let $\mathbf{S}_t = (Y_t, X_t, Z_t, U_t)$ be the joint Markov process satisfying
\cref{eq:drift_confounded_sde}, and assume that the IV exogeneity condition defined in \cref{eq:orthogonal_instruments} holds. Let $R_{t,t+\tau}$ describe the prediction residual at time $t+\tau$
\begin{equation}
 R_{t,t+\tau} = Y_{t+\tau} - Y_t - \int_t^{t+\tau}f_y(Y_s,X_s) ds .
    \label{eq:residual_def}
\end{equation}
Then the following conditional moment restrictions hold
\begin{equation}
    \frac{d}{d\tau}\,\mathbb{E}\!\left[\, Y_{t+\tau} \,\big|\, \mathcal{F}_t^z\,\right]
    \;=\; \mathbb{E}\!\left[\, f_y(Y_{t+\tau},X_{t+\tau}) \,\big|\, \mathcal{F}_t^z\,\right],
    \label{eq:residual_moment_condition_diff}
\end{equation}
and, equivalently, for every $t \geq 0$ and $\tau > 0$,
\begin{equation}
    \mathbb{E}\!\left[\, R_{t, t+\tau} \,\big|\, \mathcal{F}_t^z \,\right] \;=\; 0.
    \label{eq:residual_moment_condition}
\end{equation}
\end{proposition}
\begin{proof}
We start by following~\cite{Hansen_93} and apply Itô's formula to the test functions $\phi \in C^2(\mathbb R^{d_y+d_x+d_z+d_u}; \mathbb R)$ evaluated on the process state $\mathbf{S}_t$, yielding
\begin{equation}
    M_{t+\tau}^\phi - M_{t}^\phi \;:=\; \phi(\mathbf{S}_{t+\tau}) - \phi(\mathbf{S}_t) - \int_{t}^{t+\tau} \mathcal{L}\phi(\mathbf{S}_s)\, ds.
    \label{eq:martingale_identity}
\end{equation}
%We further assume $\int \left| \mathbf{\sigma}(\mathbf{S})^{\top} \nabla \phi(\mathbf{S}) \right|^{2} \, d\pi < \infty$, 
Here, the resulting process $M_t^\phi$ is a continuous local martingale. We choose the linear test function $\phi_j(\mathbf{S}_t) = Y_t$ such that $M_t^\phi$ becomes a true martingale under the finite first moment and linear growth assumptions. $\mathcal{L}$ denotes the generator of the full process including the unobserved state, as defined in~\cite{pavliotis2014stochastic}. We take expectations conditioned on the filtration of the instrument process at time $t$. As $\mathcal{F}_t^z \subset \mathcal{F}_t^\mathbf{S}$, the conditional expectation of any future martingale increment remains zero and thus
\begin{equation}
    \mathbb{E}\!\left[\phi_j(\mathbf{S}_{t+\tau}) \mid \mathcal{F}_t^z\right]
    - \mathbb{E}\!\left[\phi_j(\mathbf{S}_t) \mid \mathcal{F}_t^z\right]
    \;=\; \int_t^{t+\tau}\!\mathbb{E}\!\left[\mathcal{L}\phi_j(\mathbf{S}_s) \mid \mathcal{F}_t^z\right] ds.
    \label{eq:KBE_conditional_Y}
\end{equation}
By expanding the generator for the linear test function as
$\mathcal{L}\phi(\mathbf{S}_t) = f_y(Y_t,X_t) + U_t$, \cref{eq:KBE_conditional_Y} yields
\begin{equation}
    \mathbb{E}\!\left[\, Y_{t+\tau} - Y_t -\int_{t}^{t+\tau} \!\big(f_y(Y_s,X_s) + U_s\big)\,ds \,\bigg|\, \mathcal{F}_t^z\,\right] = 0 \qquad \text{for all } \tau > 0. \label{eq:general_integral_condition}
\end{equation}
The conditional expectation of the hidden confounder $U_s$ given $\mathcal{F}_t^z$ vanishes for all $s \in [t,t+\tau] $ by the exogeneity and the linearity of expectation. This directly results in~\cref{eq:residual_moment_condition}, and under standard regularity conditions we can interchange the integral and expectation to obtain the differential form in~\cref{eq:residual_moment_condition_diff}, concluding the proof.
\end{proof}
It follows that $\mathbb{E}[\Phi\, R_{t,t+\tau}]=0$ for all $\mathcal{F}_t^z$ measurable random variables $\Phi$. This implies that we can use the filtration of $Z$ to construct many valid instruments from a single continuous-time instrument process. Choosing $\Phi = g(Z_{[0,t]})$ for any functional $g \colon C([0,t], \mathbb{R}^{d_z}) \to \mathbb{R}$ of the instrument path up to time $t$, we obtain the following continuum of moments
\begin{equation}
    \mathbb{E}\big[g(Z_{[0,t]})\,R_{t,t+\tau}\big] = 0 \qquad \text{for all } g,\tau.
    \label{eq:continuum_of_moments_g}
\end{equation}
Importantly, the conditional expectation with respect to the filtration of the instrument path in~\cref{eq:residual_moment_condition} carries much more information than just the past state $Z_t$.
Although the joint process described in \cref{eq:drift_confounded_sde} is Markov, this property does not carry over to the partially observed process. 
%Moreover, the residual depends on past values of the instrument process via the filtrations of $X$ and $Y$, which we do not condition on. 
Consequently, the entire continuous instrument path up to time $t$ constitutes a valid instrument, giving rise to path-valued moment conditions.

These moment conditions rely on temporal information between observations. When those are not available, for example when only samples from the stationary distribution are accessible, we can still derive stationary moment conditions, which we describe in \cref{sec:stationary_moments}.

\section{The Linear Case: Ornstein-Uhlenbeck Processes}
In this section we show that for an \ac{OU} process the previously described moment equations can achieve point-identification of the targeted causal effect under mild conditions on the joint process.
We instantiate the general setup described in \cref{eq:drift_confounded_sde} via an \ac{OU} process with constant diffusion coefficients and a linear drift
\begin{equation}
    d \mathbf{S}_t = 
    \underbrace{
    \begin{pmatrix} 
        \beta_{yy} & \beta_{yx} & 0 & \beta_{yu} \\ 
        \beta_{xy} & \beta_{xx} & \beta_{xz} & \beta_{xu} \\ 
        0 & 0 & \beta_{zz} & 0 \\ 
        0 & 0 & 0 & \beta_{uu} 
    \end{pmatrix}
    }_{=:A}
    \mathbf{S}_t \, dt 
    + 
    \underbrace{
    \begin{pmatrix} 
        \sigma_y & 0 & 0 & 0 \\ 
        0 & \sigma_x & 0 & 0 \\ 
        0 & 0 & \sigma_z & 0 \\ 
        0 & 0 & 0 & \sigma_u 
    \end{pmatrix}
    }_{=:\boldsymbol{\sigma}}
    \begin{pmatrix} dW^y_t \\ dW^x_t \\ dW^z_t \\ dW^u_t \end{pmatrix},
    \label{eq:IV_OU_structure}
\end{equation}
where each drift coefficient is matrix-valued with dimensions $\beta_{ab} \in \mathbb R^{d_a \times d_b}$ for $a,b \in \{y,x,u,z\}$, and the diffusion coefficients are diagonal matrices $\sigma_a \in \mathbb R^{d_a \times d_a}$ for $a \in \{y,x,z,u\}$.
We assume $A$ to be Hurwitz, so that the process is stable and stationary.
Given the system described in \cref{eq:IV_OU_structure}, identification of the target function reduces to the identification of the parameters $\beta_{yx}$ and $\beta_{yy}$.

\xhdr{Simplification of moment conditions under joint Gaussianity.}
For OU processes, \cref{eq:continuum_of_moments_g} simplifies due to the joint Gaussianity of the process. As $R_{t,t+\tau}$ consists of a sum of Gaussian random variables and linear integrals over Gaussian processes, the distribution of $(R_{t,t+\tau},Z_{[0,t]})$ is also jointly Gaussian. Thus, the conditional expectation $\mathbb{E}[R_{t,t+\tau}\mid Z_{[0,t]}]$ is linear in the conditioning variables. This has two direct consequences: First, the unconditional formulation~\cref{eq:continuum_of_moments_g} can be restricted to linear functionals $g$, as any nonlinear transformation $g$ in the moment conditions does not provide any additional identifying information. 
Second, joint conditioning on multiple past states of the instrument process does not provide additional information over conditioning on each lagged $Z_{t-\delta}$ individually (\cref{app:ou_cond_gaussian}).
Consequently, in the OU setting, \cref{eq:continuum_of_moments_g} turns into a continuum of moments
% \begin{equation}
$
    \mathrm{Cov}[R_{t,t+\tau} , Z_{t-\delta}] 
    = 0_{d_y\times d_z}
$ for all $\tau > 0,\delta \geq 0$.
    % \label{eq:ou_marginal_moments}
% \end{equation}
Due to stationarity and linearity of expectation, varying both $\tau$ and $\delta$ is redundant. So we can write equivalently
%varying $\tau$ at $\delta=0$ suffices (\cref{app:ou_cond_gaussian}). Thus, \cref{eq:ou_marginal_moments} can be written equivalently as
\begin{equation}
    \mathrm{Cov}[R_{t,t+\tau}, Z_t] = 0_{d_y\times d_z} \qquad 
    \text{for all }\tau > 0.
    \label{eq:ou_moment}
\end{equation}

\xhdr{Identification.}
In an \ac{OU} process, $U_t$ also enters the drift of $X$ additively, such that the confounder's contribution to the cross-covariance dynamics between $X$ and $Z$ vanishes by exogeneity as well. Thus, differentiating \cref{eq:ou_moment} with respect to $\tau$ together with the equivalent equation for the $X$ and $Z$ cross-covariance yields a linear ODE in the covariance states $ C(\tau) = [C_{YZ}(\tau), C_{XZ}(\tau)] = [\mathbb{E}[Y_{t+\tau}Z_t^\top], \mathbb{E}[X_{t+\tau}Z_t^\top]] \in \mathbb{R}^{(d_x+d_y)\times d_z}$, in which the confounding no longer appears explicitly. The resulting system contains an additional forcing term $C_{ZZ}(\tau) \in \mathbb{R}^{d_z\times d_z}$, which can be interpreted as a control input.
\begin{equation}
    \frac{d}{d\tau}
    \begin{pmatrix}
        C_{YZ}(\tau) \\
        C_{XZ}(\tau)
    \end{pmatrix}
    =
    \underbrace{
    \begin{pmatrix}
        \beta_{yy} & \beta_{yx} \\
        \beta_{xy} & \beta_{xx}
    \end{pmatrix}
    }_{=:A_c}
    \begin{pmatrix}
        C_{YZ}(\tau) \\
        C_{XZ}(\tau)
    \end{pmatrix}
    +
    \underbrace{
    \begin{pmatrix}
        0 \\
        \beta_{xz}
    \end{pmatrix}
    }_{=:b}
    C_{ZZ}(\tau).
    \label{eq:covariance_controlled_system}
\end{equation}
Consequently, in the case of a confounded OU process, the IV moment conditions therefore amount to matching the lagged covariances dynamics between instrument and outcome described by~\cref{eq:covariance_controlled_system} to those estimated from data. With these prerequisites, we obtain the following identifiability statement.
% following proposition states under which conditions this allows for identification of the target parameters.

\begin{proposition}[Identification in the OU setting]
Consider an \ac{OU} as in~\cref{eq:IV_OU_structure} with Hurwitz drift matrix $A$ and non-degenerate instrument diffusion $\sigma_z\sigma_z^\top\succ0$ and let $a := d_x + d_y$. Further, define the controllability matrix
\begin{equation}
    \mathcal{C} = \begin{pmatrix} b & A_c b & \cdots & A_c^{a-1} b 
    \end{pmatrix} \in \mathbb{R}^{a \times a d_z}.
    \label{eq:controllability_rank}
\end{equation}
For timepoints $\boldsymbol{\tau} = (\tau_1, \ldots, \tau_a) \in (0, \tau_{max}]^a$ drawn i.i.d.\ from an absolutely continuous distribution over $(0,\tau_{max}]$, if $\mathrm{rank}(\mathcal{C}) = a$, 
the linear system obtained by integrating the first row of \cref{eq:covariance_controlled_system}
\begin{equation}
    C_{YZ}(\tau_i)-C_{YZ}(0)
    =
    \beta_{yy}\int_0^{\tau_i} C_{YZ}(s)\,ds
    +
    \beta_{yx}\int_0^{\tau_i} C_{XZ}(s)\,ds,
    \qquad i\in\{1,\ldots,a\},
    \label{eq:lin_sys_prop}
\end{equation}
almost surely has a unique solution, identifying the causal effect $\beta_{yx}$ and the self-effect $\beta_{yy}$.

\label{prop:controllability}
%\label[proposition]{prop:controllability}
\end{proposition}
\emph{Proof sketch.}
The proof (presented in detail in~\cref{proof:prop_2}) relies on three steps. Given \cref{eq:lin_sys_prop}, the proposition can directly be viewed as a rank condition on a matrix of $a$ observed timepoints of the trajectory $C(\tau)$. In LTI systems, the controllability condition \cref{eq:controllability_rank} ensures that the trajectory can span the whole state-space~\citep{bellman1970structural}. However, for a fixed control input, there may still exist initial conditions for which some modes of the system remain unobserved. Using the stationary Lyapunov equation, the proof shows that such initial conditions do not occur under a non-degenerate instrument diffusion $\sigma_z\sigma_z^\top\succ0$ almost surely.
Overall, the proposition thus states that given controllability, the causal parameters $\beta_{yx}$ and $\beta_{yy}$ are identified by fitting the covariance ODE in~\cref{eq:covariance_controlled_system} at a finite number of lags $\tau$.

\xhdr{Estimation.} We estimate the drift coefficients through a two-stage procedure based on the integrated covariance identities in~\cref{eq:lin_sys_prop}. For fixed $L>a$ lags sampled as in~\cref{prop:controllability}, define
\begin{equation}
B_\ell:=C_{YZ}(\tau_\ell)-C_{YZ}(0),
\qquad
G_\ell:=\int_0^{\tau_\ell}C(s)\,ds,
\qquad \ell=1,\ldots,L.
\end{equation}
The first stage estimates these moments from the observed trajectory. Specifically, we estimate $C_{YZ}(\tau_\ell)$ by Nadaraya--Watson kernel regression~\citep{Walk2010} of the observed cross-products $Y_{t_j}Z_{t_i}^{\top}$ on their time lags $t_j-t_i$, using a bandwidth that shrinks with sample size. Together with the empirical zero-lag covariance $\widehat C_{YZ,n}(0)=n^{-1}\sum_{i=1}^nY_{t_i}Z_{t_i}^{\top}$, this yields $\widehat B_{\ell,n}:=\widehat C_{YZ,n}(\tau_\ell)-\widehat C_{YZ,n}(0)$. We estimate $G_\ell$ directly by importance-weighted averages of lagged cross-products involving both $Y$ and $X$, correcting for the possibly nonuniform lag density induced by the known observation-gap distribution.

The second stage fits the empirical moment equations by least squares. Stacking $\widehat B_n=(\widehat B_{1,n},\ldots,\widehat B_{L,n})$ and $\widehat G_n=(\widehat G_{1,n},\ldots,\widehat G_{L,n})$, the estimator admits the closed-form expression
\begin{equation}
(\widehat\beta_{yy,n},\widehat\beta_{yx,n})
:=\widehat B_n\widehat G_n^\dagger
\in\operatorname*{arg\,min}_{\Theta\in\mathbb R^{d_y\times a}}
\left\|\widehat B_n-\Theta\widehat G_n\right\|_F^2,
\label{eq:integral-iv-estimator}
\end{equation}
where ${}^\dagger$ denotes the Moore--Penrose pseudoinverse, selecting the minimum-Frobenius-norm solution when the minimizer is nonunique. This combines nonparametric covariance estimation with a single linear least-squares solve for the causal drift coefficients.
All details of the construction, a differential version of the estimator that performs similarly in experiments, and the proof of the following consistency statement are in~\cref{app:estimator_discussion}.

\begin{restatable}[Consistent estimator in the OU setting]{theorem}{consistencyint}
\label{thm:draft-integral-consistency}
In the setting of \cref{prop:controllability}, the estimator in \cref{eq:draft-integral-ls-estimator} is consistent, i.e.,
% \begin{equation}
$\widehat{\beta}_{yy,n}\overset{p}{\rightarrow}\beta_{yy}$ and $\widehat{\beta}_{yx,n}\overset{p}{\rightarrow}\beta_{yx}$ as $n\to\infty$.
% \end{equation}
\end{restatable}
In summary, we have established identifiability and developed a consistent estimator for the linear (OU) continuous-time IV setting. Before demonstrating the practical efficacy of our estimator empirically, we also develop a tractable estimation procedure for the generically unidentifiable and complex nonlinear case.

\section{The Nonlinear Case}
In the \ac{OU} setting, identification is governed by a controllability condition, which can ultimately be tested for from observed data.
In the nonlinear case, identification and practical estimation become much more challenging, as linear functionals $g$ do no longer exhaust the identifying information contained in the moment equation~\cref{eq:residual_moment_condition_diff} and furthermore, the target parameters generally cannot be moved out of the expectation, as in~\cref{eq:covariance_controlled_system}. 

\xhdr{Identification.} It is known from the static case, that nonparametric instrumental variable estimation, which seeks to recover a nonlinear function from conditional moment conditions, is an ill-posed problem~\citep{bennett2023minimaxinstrumentalvariableregression, zhang2023instrumentalvariableregressionkernel, shen2026instrumentalvariableanalysisstructural, vanderlaan2026nonparametricinstrumentalvariableinference}. 
%The reason for this is that the variation of the instrument often does not suffice to uniquely pin down the nonlinear target function. Furthermore, the moment conditions can depend on the parameters discontinuously. 
In the nonlinear setting, the goal is to identify the parameters $\theta \in \mathbb R^p$ parameterizing the target drift function $f_{y;\theta} \in \mathcal{H}$ in the general SDE model described in \cref{eq:drift_confounded_sde}. Generally, $\theta$ is unidentifiable for flexible function classes $\mathcal{H}$, i.e., the conditional moment restrictions in~\cref{eq:residual_moment_condition} are satisfied for multiple parameter settings. The standard remedy from the literature~\citep{bennett2023minimaxinstrumentalvariableregression} is to regularize the problem to obtain one compatible candidate $\theta$ in the identified set, for instance, aiming at the least norm solution.
We follow this approach in our continuous-time setting.

\xhdr{Estimation.} We start from the unconditional form of the previously derived moment condition $ \mathbb{E}\big[g(Z_{[0,t]})\,R_{t,t+\tau;\theta}\big] = 0$ for all $g,\tau$, where $R_{t,t+\tau;\theta}$ now depends on the drift function $f_{y;\theta}$ via~\cref{eq:residual_def}. We evaluate the residual at a fixed time horizon $\tau = h$, where $h$ is a hyperparameter. For a moderate choice of $h$, the signal to noise ratio of the drift contribution increases. At the same time, the correlation to the past instrument path typically decreases as $h$ increases. When the sampling frequency is sufficiently high, the integral terms in $R_{t,t+\tau}$ can be approximated using standard quadrature schemes. Otherwise, simulation-based estimation methods are required. 

We can then formulate the nonlinear estimation problem as a minimax problem of the risk $\mathcal{R}_z$ associated with the moment conditions~\cref{eq:residual_moment_condition} \citep{bennett2023minimaxinstrumentalvariableregression, ailer2025targetedsequentialindirectexperiment}. Concretely, we minimize the parameters of the target function with respect to the worst case functional $g$, where we assume that $g$ lies in the reproducing kernel Hilbert space (RKHS) $\mathcal{G}$ induced by some characteristic kernel $k_z: C([s,t], \mathbb{R}^{d}) \times C([s',t'], \mathbb{R}^{d}) \to \mathbb{R}$. Together with the regularization term $\Omega:  \mathbb R^p \to \mathbb{R}$ this gives the following dual optimization problem
\begin{equation}
\theta^\star
=
\arg\min_{\theta} \mathcal{R}_z
=
\arg\min_{\theta}
\Bigg[
\sup_{g \in \mathcal{G},\;\|g\|_{\mathcal{G}} \le 1}
\left(
\mathbb{E}\left[
R_{t,t+h;\theta}\, g(Z_{[0,t]})
\right]
\right)^2
+
\lambda \Omega(\theta)
\Bigg].
\label{eq:objective_nonlinear}
\end{equation}
Following~\cite{zhang2023instrumentalvariableregressionkernel}, we assume that $g$ lies within a unit ball of its RKHS $\mathcal G$. This allows us to avoid a computationally expensive explicit dual optimization, as the inner supremum admits an analytical solution, so that we obtain the risk
\begin{equation}
\mathcal{R}_z(\theta)
= \mathbb{E}_{i,j}\left[
R^i_{t,t+h;\theta}\, R^j_{t,t+h;\theta}\,
k_z\big(Z^i_{[0,t]}, Z^j_{[0,t]}\big)
\right] + \lambda \Omega(\theta).
\end{equation}
Using the V-statistic to empirically estimate the risk, we finally get
\begin{equation}
\widehat{\mathcal{R}}_z(\theta)
= \frac{1}{n^2} \sum_{i=1}^{n} \sum_{j=1}^{n}
R^i_{t,t+h;\theta}\,R^j_{t,t+h;\theta}\,
k_z\!\big(Z^i_{[0,t]}, Z^j_{[0,t]}\big) + \lambda \Omega(\theta)
= \frac{1}{n^2} \mathbf{r}^\top K_z\, \mathbf{r} + \lambda \Omega(\theta),\label{eq:finalobjective}
\end{equation}
where $K_z \in \mathbb R^{n\times n}$ describes the kernel Gram matrix, $\mathbf r$ the residual vector, and $n$ the number of sampled path sections. 

\xhdr{Signature kernel.} The risk $\mathcal{R}_z$ assumes access to a characteristic kernel on path spaces to express the functionals of the instrument paths in the RKHS $\mathcal G$. We propose using the signature kernel~\citep{manten2025signaturekernelconditionalindependence,Salvi_2021,lee2023signaturekernel}, which has gained increasing popularity in recent years for various types of sequential data. More details are found in~\cref{app:sig_kernel}. 
%While the estimation method presented in this section resembles static nonparametric instrumental variable methods, the continuous-time setting introduces multiple challenges that are fundamentally distinct, as the instruments are path-valued, and the integration of the candidate drift over the interval $h$ introduces additional smoothing.
\section{Experiments}
\xhdr{\ac{OU} setup and baselines.}
We evaluate our method on simulated \ac{OU} process trajectories across different systems. For each system, we randomly sample drift and diffusion matrices with sparsity according to \cref{eq:IV_OU_structure}, and all nonzero parameters bounded away from zero. The observation timepoints are irregularly sampled with uniformly distributed observation intervals, detail are described in~\cref{app:exp_details}.

We compare our method to parameter estimates obtained from naive Maximum Likelihood estimates~\citep{pavliotis2014stochastic} ignoring the unobserved state $U$. Due to a lack of continuous-time IV baselines, we also compare to the $\mathrm{VAR}(1)$ estimator presented in~\cite{thams2024} working with the same number of regularly sampled data from the same trajectory. Additionally, in~\cref{app:exp_details}, we compare to the discrete-time estimator for the normalized parameters in~\citep{mogensen2023}, where we construct the ratio estimates from both individually estimated variables $\beta_{yx}$ and $\beta_{yy}$. 
%We highlight that there exist no other IV estimation techniques in dynamical systems.

\begin{figure}
    \centering
    \begin{subfigure}{0.455\linewidth}
        \centering
        \includegraphics[height=3.5cm]{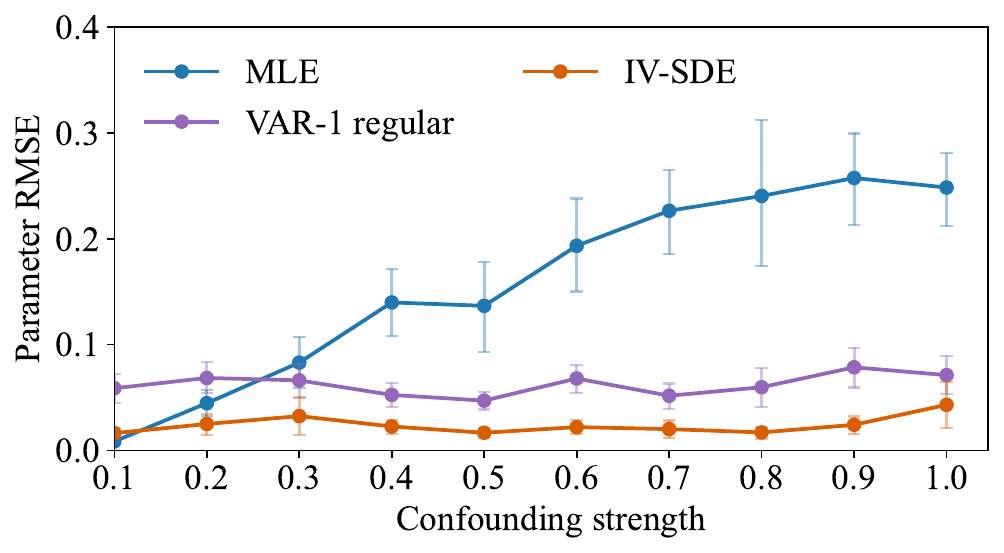}
    \end{subfigure}
    \hfill
    \begin{subfigure}{0.265\linewidth}
        \centering
        \includegraphics[height=3.5cm]{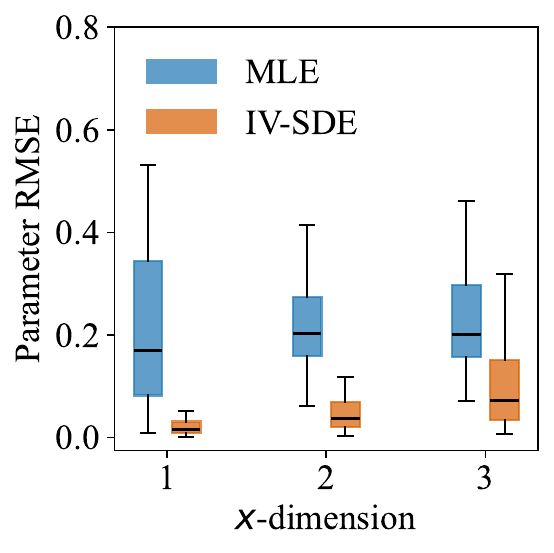}
    \end{subfigure}
    \hfill
    \begin{subfigure}{0.265\linewidth}
        \centering
        \includegraphics[height=3.5cm]{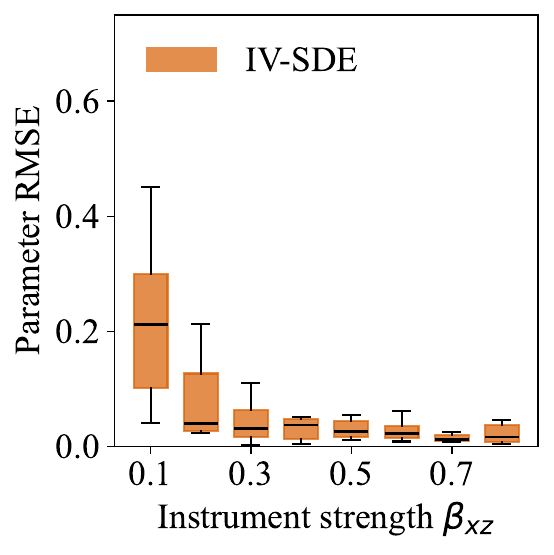}
    \end{subfigure}
    \vspace{-4mm}
    \caption{RMSE of $\beta_{yy}$ and $\beta_{yx}$ for a confounded OU process (one instrumental and one outcome process throughout, except in (b)). \textbf{Left:} Estimation error under varying confounding strength. \textbf{Center:} Recovery of multidimensional causal effects from a single instrument process. \textbf{Right:} Estimation error as a function of instrument–covariate coupling strength $\beta_{xz}$. Error bars indicate 95\% confidence intervals, boxplots show medians and interquartile ranges. Each evaluation is based on 20 systems (left) and 10 systems (middle/right).}
    \label{fig:combined_all}
    \vspace{-5mm}
\end{figure}

\xhdr{IVSDE identifies causal effects under confounding.} \Cref{fig:combined_all}(left) shows the parameter estimation error over both $\beta_{yy}$ and $\beta_{yx}$ in a confounded OU process, with increasing confounding strength through parameters $\beta_{xu}$ and $\beta_{yu}$. The \ac{MLE} estimate of the parameters is clearly increasingly biased under growing confounding, reflecting that MLE cannot distinguish between the confounding effect and the causal influence of process $X$ on $Y$.
By contrast, the continuous-time IV estimate remains stable across confounding levels. Given the same number of samples on a regular grid, discrete IV methods for time series remain biased as they assume the IV sparsity in the finite-time transition maps instead of on the infinitesimal level, as explained in \cref{app:discrete_vs_conti_independencies}. 

\xhdr{The proposed method identifies multi-dimensional causal effects with a single instrument.} 
Unlike static IV methods, which require at least as many instruments as causal effects for identification in the linear setting, a single instrumental process in continuous-time can identify multi-dimensional causal effects. \Cref{fig:combined_all} (middle) shows the parameter estimation error depending on the dimensionality of the covariate process given a single instrument process. The total number of estimated parameters is always $d_x+1$, accounting for both $\beta_{yx}$ and $\beta_{yy}$. The results reveal that the proposed IV method may even be used to estimate four-dimensional causal effects using a single instrumental process and still obtains a lower average error than MLE. However, the variance of the IV estimator increases with growing dimensionality of the targeted causal effect.

\xhdr{Influence of instrument strength.} The proposed IV method relies on the coupling between instrument and covariate process, which is also captured by the relevance assumption. \Cref{fig:combined_all} (right) shows that the mean estimation error and variance decrease with stronger coupling parameter $\beta_{xz}$. This is consistent with the intuition that larger coupling parameters induce more exogenously generated variation in the covariate process, thus providing a stronger instrument.

  \begin{wraptable}{r}{0.3\textwidth}
      \centering
      \vspace{-5mm}
      \caption{Parameter RMSE
    on the diabetes dataset.}
      \label{tab:diabetes-small}
      \vspace{-2mm}
      \resizebox{\linewidth}{!}{%
          \begin{tabular}{lr}
              \toprule
              Method & RMSE \\
              \midrule
              IV-SDE & \textbf{0.0157} $\;{\scriptstyle(0.0035)}$ \\
              MLE & 0.0998 $\;{\scriptstyle(0.0003)}$ \\
              \bottomrule
          \end{tabular}%
      }
      \vspace{-5mm}
  \end{wraptable}
  
\begin{wrapfigure}{r}{0.5\columnwidth}
    \centering
    \vspace{-4mm}
    % \begin{subfigure}{0.48\linewidth}
    %     \centering
        \includegraphics[height=3.42cm]{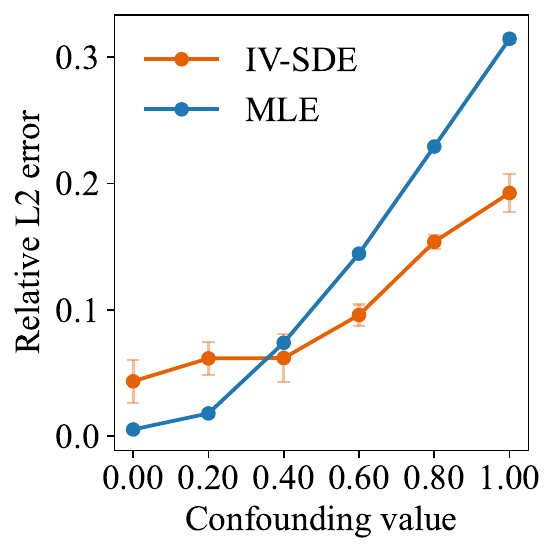}
        % \label{fig:mse_drift}
    % \end{subfigure}
    % \hspace{0pt}
    % \begin{subfigure}{0.48\linewidth}
    %     \centering
        \includegraphics[height=3.42cm]{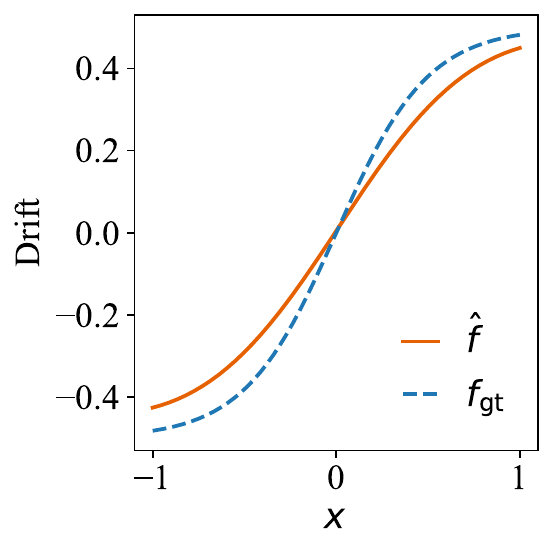}
        % \label{fig:sigmoid_sde}
    % \end{subfigure}
    \vspace{-2mm}
    \caption{Results for a nonlinear drift parametrization. (left) Relative drift RMSE over the $X_t,Y_t$ distribution. (right) Ground truth sigmoidal drift $f_{\mathrm{gt}}$ and learned drift $\hat{f}$ parameterized with an MLP.}
    \label{fig:nonlinear_results}
    \vspace{-5mm}
\end{wrapfigure}

\xhdr{Kernel-based estimation of confounded nonlinear effects.}
In the nonlinear setting, inspired by~\cite{saengkyongam2022exploitingindependentinstrumentsidentification}, we set up the drift $f_y$ as $f_y(Y_t,X_t) = \alpha Y_t + \beta X_t + \sum_{i=1}^p w_i e^{-(X_t-c_i)^2/\sigma_{\mathrm{rbf}}^2} - \sum_{i=1}^p w_i e^{-(X_t+c_i)^2/\sigma_{\mathrm{rbf}}^2}$, where $w\in \mathbb R^p$ describes a randomly sampled parameter vector of radial-basis function weights and $c\in \mathbb R^p$ their centers.
%We construct $f_y$ to be odd, such that $Y$ remains a zero-mean process. 
We parameterize the target function $f_{y;\theta}$ with a basis function expansion containing the same radial basis function centers and variances, as well as two linear terms for $X$ and $Y$.
%We assume a three-dimensional instrument process and a one-dimensional covariate process. The confounding noise follows a double-well SDE. For comparison, we use an MLE-like drift estimator based on a Gaussian quasi-likelihood under Euler–Maruyama discretization, which uses the same radial-basis function expansion for the drift, a more detailed of the setup description can be found in~\cref{app:exp_details}. 
\Cref{fig:nonlinear_results} (left) clearly shows that the IV-based estimator outperforms the MLE estimation for larger confounding values, while the MLE performs better for small and no confounding. This behavior of these estimation methods can be explained by two observations. First, IV methods are inherently less efficient than MLE, since they only exploit the variation explained by the instrument instead of the full likelihood. Furthermore, the signature kernel-based optimization introduces a more complex optimization landscape, which might contribute to the lower performance in settings where MLE is only slightly biased. With higher confounding, however, this is outweighed by the bias of the MLE estimate.

\xhdr{Neural parametrization.}
We can also use a small MLP to parameterize the class of the target drift function $f_y$. In this setting, we assume the ground truth drift function to be linear in $Y$ and use a sigmoid function to describe the dependence on the covariates $X$. \Cref{fig:nonlinear_results} (right) shows that we recover the qualitative shape of the target function.

\xhdr{Diabetes dataset.} We setup a semi-synthetic experiment motivated by pharmacokinetic/pharmacodynamic models for type 2 diabetes~\citep{Subramanian_2024, Naik_2026}. We aim to identify the patient’s insulin sensitivity, which describes how their glucose level responds to insulin and is the main physiological factor impaired by the disease. Both glucose and insulin levels are confounded by unobserved gut glucose levels, and external insulin dosing schedule is used as an instrument. The resulting drift is modeled as $g_y(\mathbf{S}) :=f_y(Y_t,X_t)+c_{yu} U_t = 
c_{yu}U_t-a_y Y_t-\gamma_{xy}X_tY_t$.
\Cref{tab:diabetes-small} shows that compared to an MLE-based parameter fit, our nonlinear estimator retrieves all function parameters with lower RMSE across five runs. More runs including parameter ablations are in~\cref{app:exp_details}.

\vspace{-3mm}
\section{Conclusion}
\vspace{-2mm}
We introduced a framework for instrumental variable estimation in continuous-time stochastic processes governed by SDEs by deriving path-based moment conditions resulting from the structural IV assumptions. For OU processes, we established a controllability-based identification condition together with a consistent estimator. For nonlinear effects, we proposed a minimax estimation scheme based on path-valued moment conditions. We have shown the empirical performance on synthetic data and proposed SDE simulators for diabetes from the literature. Our results also verify that a single instrument process indeed carries enough information to recover multi-dimensional effects.

We believe that with the currently growing interest in learning continuous-time dynamics~\citep{kostic2024learninginfinitesimalgeneratorstochastic,lorch_sdes} the causal interpretation of these models is becoming increasingly relevant. Our results also highlight once more that the transition from discrete time to continuous time is often not trivial as local structure does often not transfer to finite-step transitions. Across our simulated systems, including the glucose--insulin example, respecting this structure improves estimation.
Interesting direction for future work include a better understanding of identifiability in the nonlinear case and uncertainty quantification that remain reliable with weak instruments. Also, theoretical extensions beyond stationarity would be interesting, albeit stationarity does not appear to be mandatory for practical estimation.
% The connection to controllability also suggests a direction for experimental design: jointly choosing instrument trajectories and observation times to make causal effects easier to identify.

\subsection*{AI use statement}
In this work, we used generative AI tools to help generate synthetic data sets, support formulating mathematical claims, assist in the writing of proofs, propose and refine hypotheses, design or provide feedback on research methodology and experiments, support with the implementation of methods and support qualitative and thematic data analysis. We have not used generative AI tools to provide critical ingredients for proving mathematical claims, help develop conceptual frameworks, interpret results and assist with translation. Cleaning and reformatting datasets is not applicable to this work. 

While we used AI in the research process, the final content of this paper was manually constructed, checked and verified with care. We have manually reviewed all LLM written code for functionality and correctness. We take responsibility for the final content of this work, including text, claims and artifacts produced with the aid of generative AI.

\subsection*{Reproducibility statement}

We prove all main mathematical statements in the appendix and the main paper itself and precisely refer to previous methods, when necessary. Further, we provide a thorough discussion of the method and its main assumptions in the appendix. We supply full descriptions of all synthetically generated datasets and used methods in the appendix. We provide the main methods in an anonymized form as source code at time of submission, the repository with full reproducible result plots will be available upon acceptance.

\bibliography{iclr2027_conference}
\bibliographystyle{iclr2027_conference}

\appendix

\newpage

\section{Extended Setup Discussion}
\label{app:setup}

Our setup relies on the standard instrumental variable assumptions, and it is worth discussing them in more detail and disentangling them from other assumption made throughout this paper.

\subsection{Static IV assumptions}
\begin{figure}
    \centering
      \begin{tikzpicture}
     \node[obs] (X1) {$Z$};
     \node[obs, right =of X1] (X2) {$X$};
     \node[obs, right =of X2] (X3) {$Y$};
     \node[latent, above = of $(X2)!0.5!(X3)$] (X4) {$U$};
     \edge{X1}{X2}
     \edge[color=MidnightBlue]{X2}{X3}
     % \edge[out=195, in=-15]{X3}{X2}
     \edge{X4}{X2,X3}
     % \draw[->, >={triangle 45}, thick] (X1) edge[loop below, looseness=8] (X1);
     % \draw[->, >={triangle 45}, thick] (X2) edge[loop below, looseness=8] (X2);
     % \draw[->, >={triangle 45}, thick, color=MidnightBlue] (X3) edge[loop below, looseness=8] (X3);
     % \draw[->, >={triangle 45}, thick] (X4) edge[loop above, looseness=8] (X4);
  \end{tikzpicture}
    % \begin{tikzpicture}[
    %     node distance=2.2cm,
    %     observed/.style={
    %         circle,
    %         draw,
    %         minimum size=9mm,
    %         inner sep=0pt
    %     },
    %     latent/.style={
    %         circle,
    %         draw,
    %         dashed,
    %         minimum size=9mm,
    %         inner sep=0pt
    %     },
    %     causal/.style={
    %         -{Latex[length=2mm]},
    %         thick
    %     }
    % ]
    %     \node[observed] (Z) {\(Z\)};
    %     \node[observed, right=of Z] (X) {\(X\)};
    %     \node[observed, right=of X] (Y) {\(Y\)};
    %     \node[latent, above=1.5cm of X] (U) {\(U\)};

    %     \draw[causal] (Z) -- (X);
    %     \draw[causal] (X) -- (Y);
    %     \draw[causal] (U) -- (X);
    %     \draw[causal] (U) -- (Y);
    % \end{tikzpicture}
    \caption{Standard instrumental-variable causal graph.
    The instrument $Z$ affects the outcome $Y$ only through the
    treatment $X$, while the latent variable $U$ confounds the
    relationship between $X$ and $Y$. The causal effect of interest is the direct dependence of $Y$ on $X$ highlighted in blue.}
    \label{fig:standard_iv_scm}
\end{figure}
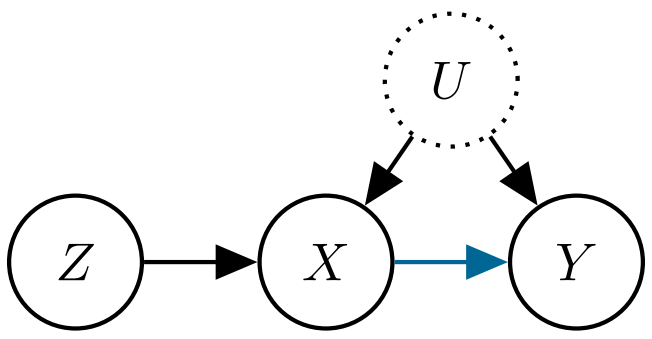

The instrumental variable setting itself is defined by the three conditions \textit{relevance, exclusion and exogeneity}~\citep{hernan2020causal, Pearl_causality}. Relevance requires that $X$ is not independent of $Z$, the exclusion restriction requires that $Z$ only affects $Y$ through $X$ and exogeneity requires that $U$ is independent of $Z$. A simple \textit{structural causal model} fulfilling these instrumental variable conditions is found in~\cref{fig:standard_iv_scm}. Relaxing any of these conditions is known to yield impossibility statements without strong replacements such as sign- or subset validity restrictions. For such very specific, IV-related settings, one can provably only obtain bounds or sensitivity regions and no point-identification~\citep{masten_poirier, balke}. While the literature contains ample discussion of the IV assumptions, they have not impaired the practical relevance and usage of IV estimation.

In addition to these three defining conditions, IV methods require a fourth assumption for identification~\cite{hernan2020causal}, which usually takes the form of the additivity assumption. This assumption implies that the confounder only enters the outcome variable in an additive way. This assumption is substantive, but it is just as well standard and provably necessary: If this additive separability is not assumed the effect is unidentifiable unless other strong assumptions are introduced. Such alternatives include adding monotone control functions or shape restrictions \citep{Imbens,chesher} or give up on identifiability and analyze partial identification.

% Figure

\subsection{Transfer to the SDE setting}
We transfer the previously described defining assumptions of the IV setting together with the necessary additivity assumption to the continuous-time dynamic setting. We make the standard assumption of a a diagonal diffusion matrix with coordinate-wise state-dependent entries. Consequently, we assume all causal effects between the variables to be encoded in the drift functions. We discuss a relaxation of this in a later section.

We encode the structural IV assumptions in the sparse interaction graph of the SDE.
Thus, the \textbf{relevance} condition requires the drift of $X$ to depend on the instrument $Z$. Similarly, the \textbf{exclusion} restriction requires that $Z$ does not enter the drift of outcome process $Y$ and only influences $Y$ through process $X$. \textbf{Exogeneity} is best expressed through the path-filtration condition $\mathbb E(U_s|\mathcal F_t^z)=0,\quad s \geq t$. The \textbf{Additivity} assumption can be directly transferred to the dynamic setting by requiring additivity of the unobserved state $U_t$ in the drift function of process $Y$.

Without the assumption of a diagonal, component-wise state-dependent diffusion, other definitions of the IV setup would also be possible in continuous-time, for which our proposed moment conditions would remain valid.

In its most general form, the moment condition presented in \cref{prop:moment_conditions} can be written as 
\begin{equation}
    \mathbb E [R_{t,t+h} | \mathcal{F}_t^Z] = \int_t^{t+h}\mathbb E[U_s|\mathcal F_t^Z]ds + \mathbb E[M_{t+h}^Y-M_t^Y | \mathcal F_t^Z].
\end{equation}
\begin{equation}
M_t^Y = 
\int_0^t\sigma_{yy}(S_s)dW_s^Y
+\int_0^t\sigma_{yx}(S_s)dW_s^X
+\int_0^t\sigma_{yz}(S_s)dW_s^Z
+\int_0^t\sigma_{yu}(S_s)dW_s^U.
\end{equation}

Hence, the moment restrictions remain valid as long as $M_t$ is a martingale and the exogeneity of the instrument holds. In case of an arbitrary non-diagonal and state-dependent diffusion matrix, $M_t$ remains a martingale. However, a non-diagonal diffusion matrix may break the validity of the moments through a violation of the exogeneity condition. This may occur in two ways: 

First, exogeneity may be violated if $Z$ and $U$ are directly mixed by sharing a mutual innovation, specifically $
\bigl(\sigma(S_t)\sigma(S_t)^\top\bigr)_{zu} \neq 0$. This introduces shared information between the history of $Z$ and the history of $U$, which through persistence in the latent process also allows for a prediction of future $U$. Second, exogeneity may also not hold, if the diffusion coefficient of either process depends on the state of the other described by $\sigma_u=\sigma_u(U_t,Z_t)$ or $\sigma_z=\sigma_z(Z_t,U_t)$.

We note that, as long as exogeneity is fulfilled, diffusion mixing as well as state dependent diffusions may introduce proper confounding bias in the naive MLE estimator even without drift confounding, while the moment conditions in \cref{prop:moment_conditions} and thus the proposed estimation remain valid. Consequently, we may relax the assumption on a diagonal diffusion matrix by only requiring that exogenity holds, and could also define a more general continuous-time IV setting than described by the SDE system in~\cref{eq:drift_confounded_sde}. 

We decide to present this paper with the more restrictive drift confounding as in \cref{eq:drift_confounded_sde}, because we argue that proper causal confounding is intuitively conveyed via the drift instead of shared diffusion noise or state-dependent volatility. We discuss the influence of a non-diagonal diffusion matrix on our proposed methods in~\cref{app:non_diag_diffusion_matrix}.

\subsection{Discussion of further assumptions.}
Beyond the previously discussed conditions defining the IV setting, we make the following further assumptions on the process:

\xhdr{Zero-mean.} We assume the joint process to have zero-mean. This holds without loss of generality, as the observed parts of the process can be centered with the observed mean. The moment conditions in both the linear and nonlinear setup retrieve the function $f_y(Y_t,X_t)$ up to translation, as we can only identify the sum of the intercept of $f_y(Y_t,X_t)$ and the unobserved $\mu_u = \mathbb E(U_t)$. However, for prediction of process $Y$ under intervention on process $X$ this translation cancels and the total outcome drift of process $Y$ remains unchanged. Hence, we can assume $\mathbb{E}[U_t]=0$ without loss of generality. As presented, the current model setup requires $d_u = d_y$. We interpret $U$ as the effective latent contribution instead of necessarily the full underlying state. Its dynamics may also depend on additional unobserved variables, the proposed moment conditions will remain valid as long as exogeneity is fulfilled.

\xhdr{Stationarity.} We assume the joint process to be stationary, the non-stationary setting is only adjacent to our setup. We make use of stationarity for identification and estimation from a single curve. Together with the assumption of ergodicity, we can estimate the moments using time averages from a single trajectory. 
In the nonstationary setting, the moments would generally depend on the absolute observation time. Thus, they could not be pooled along single trajectories without introducing additional assumptions. Estimation would instead require e.g. multiple independent trajectories or assumptions on the local stationarity of the process. We leave this extension to future work.

\section{Distinction from Discrete-Time Models}
\label{app:discrete_vs_conti_independencies}
In the following section we demonstrate the difference between structural assumptions in discrete-time models and continuous-time models as also described in \cite{Mogensen2018CausalLF}. We specifically show that local IV structure on an infinitesimal level does not imply IV sparsity in finite interval transition maps in discrete-time models.

Consider as an example a general \ac{OU} process with states $\mathbf{S}_t$ with drift matrix $A$ and diagonal diffusion matrix $\boldsymbol{\sigma}$. 
When all components of $\mathbf{S}_t$ are observed on an equidistant grid with sampling interval $h > 0$, \cite{bergstrom} showed that the discrete observations satisfy the $\mathrm{VAR}(1)$ process with transition matrix $\Phi$:
\begin{equation}
    \mathbf{S}_{t} \;=\; e^{Ah}\, \mathbf{S}_{t-h} \;+\; \eta_t^{h} = \Phi \mathbf{S}_{t-h}+ \eta_t^h,
    \qquad t = h, 2h, \dots,
    \label{eq:var1-representation}
\end{equation}
Here, the noise follows:
\begin{equation}
    \eta_t^{(h)} \;=\; \int_{t-h}^{t} e^{A(t-s)}\, \boldsymbol{\sigma}\, dW_s,
\end{equation}
thus satisfying
\begin{equation}
    \mathbb{E}[\eta_t^{(h)}] = 0,
    \qquad
    \mathbb{E}\!\left[\eta_t^{(h)} (\eta_t^{(h)})^\top\right] \;=\; \int_0^{h} e^{As}\, \boldsymbol{\sigma}\boldsymbol{\sigma}^\top\, e^{A^\top s}\, ds
\end{equation}
and $\mathbb{E}[\eta_t^{(h)} (\eta_{t-r}^{(h)})^\top] = 0$ for all $r \geq h$. 

The discrete-time transition matrix $\Phi = e^{Ah}$, however, generally does not maintain the sparsity pattern of drift matrix $A$. Therefore, the local independencies in $A$ are typically not visible in the discrete $\mathrm{VAR}(1)$ representation~\cite{Mogensen2018CausalLF}.

Specifically, for the OU process with IV structure described in \cref{eq:IV_OU_structure} with states $\mathbf{S}_t = (Y_t,X_t,Z_t,U_t)$, the IV sparsity encoded in the drift (\cref{eq:IV_OU_structure}) does not carry over to the transition matrix in the $\mathrm{VAR}(1)$ model of the process.
While the block-diagonal structure of $A$ ensures that the exogeneity of the instrument and confounding process are preserved in the finite-time transition matrix, the exclusion restriction does not hold. Specifically, $A_{yz}=0$ does not imply $\Phi_{yz} = 0$.
For a concrete example, consider
\begin{equation}
A =
\begin{pmatrix}
-1.4 & -1 & 0 & -0.5 \\
-0.25 & -1.2 & -1 & -0.5 \\
0 & 0 & -1 & 0 \\
0 & 0 & 0 & -1
\end{pmatrix},
\end{equation}
but for $h=1$
\begin{equation}
\Phi = e^{Ah} \approx
\begin{pmatrix}
0.2803 & -0.2845 & 0.1541 & -0.0806 \\
-0.0711 & 0.3372 & -0.3462 & -0.1538 \\
0 & 0 & 0.3679 & 0 \\
0 & 0 & 0 & 0.3679
\end{pmatrix},
\end{equation}
where $\Phi_{yz}\neq 0$ despite $A_{yz}=0$.

Consequently, imposing a local IV structure in a continuous-time model implies a different assumption about the underlying mechanism than enforcing IV sparsity in discrete-time models.

\section{Stationarity moment conditions}\label{sec:stationary_moments}

In the setting where only observations from the stationary distribution are accessible, we can still setup moment conditions: since $\mathbb{E}[\mathcal{L}\phi(\mathbf{S}_t)] = 0$ (presented by~\citealp{Hansen_93}), we can directly make use of the stationary covariance between any function of the instrument state $Z_t$ and the outcome $Y_t$. For this, we choose the test function as $\phi(\mathbf{S}_t) = g(Z_t)Y_t$, where $g(\cdot)$ is a projection of the instrument state. Expanding the generator expression and using the exogeneity of the instrument, this yields
\begin{equation}
    \mathbb{E}\Big[ g(Z_t) (f_{y}(Y_t,X_t))  \Big] = -\mathbb{E}\left[ Y_t \left( f_z(Z_t)g'(Z_t) + \frac{1}{2}\sigma_z^2(Z_t)g''(Z_t) \right) \right].
    \label{eq:stat_confounded_final}
\end{equation}
While this moment condition can be used for estimation and may even identify the target function by using different projections $g$ under a sufficiently rich instrument process, it also assumes knowledge of the instrument dynamics in the form of the instrument drift and diffusion. We therefore only focus on the previously introduced moment conditions in the presentation of this paper.

\section{Simplification of Moment Conditions under Joint Gaussianity}
\label{app:ou_cond_gaussian}
% The conditional moment restriction cannot be extended to higher moments, if the diffusion is state dependent.
The following section briefly shows that the moment restrictions conditioned on the full instrument path reduce to moment restrictions conditioned on the individual instrument states in an OU system.
Since the OU process is jointly Gaussian, the vector $(R_{t,t+\tau}, Z_{s_1}, \ldots, Z_{s_k})$ is jointly Gaussian with zero mean for any $s_1, \ldots, s_k \leq t$. The conditional expectation of a jointly Gaussian vector is linear in the conditioning variables, so: 
\begin{equation}
    \mathbb{E}\big[R_{t,t+\tau} \mid Z_{s_1}, \ldots, Z_{s_k}\big] 
    = \mathrm{Cov}[R_{t,t+\tau}, \mathbf{Z}]\, \mathrm{Cov}[\mathbf{Z}, \mathbf{Z}]^{-1} \mathbf{Z},
\end{equation}
where $\mathbf{Z} = (Z_{s_1}, \ldots, Z_{s_k})$. Since $\mathrm{Cov}[R_{t,t+\tau}, Z_s] = 0$ for all $s \leq t$, the path-valued moment conditions reduce to moment conditions based on all individual variables $Z_s$.

We further show that the continuum of moments over $\delta$ and $\tau$ may be reduced to the single-parameter family in \cref{eq:ou_moment}. Starting from just the continuum over $\tau$

\begin{equation}
    \mathrm{Cov}\big[R_{t,t+\tau}, Z_t\big]
    = 0 \qquad \text{for all }\tau > 0.
\end{equation}

we define $\Delta=\delta+\tau$ and write, using linearity of expectation

\begin{equation}
    \mathrm{Cov}\big[R_{t,t+\Delta}, Z_t\big]
    = \mathrm{Cov}\big[R_{t,t+\delta}, Z_t\big]
    + \mathrm{Cov}\big[R_{t+\delta,t+\delta+\tau}, Z_t\big]
    = 0 \qquad \text{for all }\delta,\tau > 0.
\end{equation}

Evaluating the first term to zero and using stationarity yields:
\begin{equation}
    \mathrm{Cov}\big[R_{t,t+\tau}, Z_{t-\delta}\big]
    = 0 \qquad \text{for all }\tau,\delta > 0
\end{equation}
which is equivalent to the continuum of moments over both $\tau$ and $\delta$.

\section{Proof for \texorpdfstring{\cref{prop:controllability}}{Proposition~\ref{prop:controllability}}}
\label{proof:prop_2}
The proof for \cref{prop:controllability} consists of three steps.

\subsection{Step 1: reduction to rank condition}
Given is the ODE system with states $ C(\tau) = (C_{YZ}(\tau), C_{XZ}(\tau)) \in \mathbb{R}^{(d_y+d_x)\times d_z}$ and forcing term $C_{ZZ}(\tau) \in \mathbb{R}^{d_z\times d_z}$:
\begin{equation}
    \frac{d}{d\tau}
    \begin{pmatrix}
        C_{YZ}(\tau) \\
        C_{XZ}(\tau)
    \end{pmatrix}
    =
    \underbrace{
    \begin{pmatrix}
        \beta_{yy} & \beta_{yx} \\
        \beta_{xy} & \beta_{xx}
    \end{pmatrix}
    }_{=:A_c}
    \begin{pmatrix}
        C_{YZ}(\tau) \\
        C_{XZ}(\tau)
    \end{pmatrix}
    +
    \underbrace{
    \begin{pmatrix}
        0 \\
        \beta_{xz}
    \end{pmatrix}
    }_{=:b}
    C_{ZZ}(\tau), \quad C(0) = c_0
\end{equation}

Furthermore, sampling $a:= d_x+d_y$ random lags $\tau\geq0$, integrating the first row of the covariance ODE, and stacking the equations results in the linear system described in \cref{eq:lin_sys_prop}
\begin{equation}
K\,G(\tau) \;=\; B(\tau).
\label{eq:linsys}
\end{equation}
where \[
G(\boldsymbol{\tau})
:=
\left(
    \int_0^{\tau_1} C(s)\,ds,\,
    \ldots,\,
    \int_0^{\tau_a} C(s)\,ds
\right)
\in \mathbb{R}^{a \times a d_Z},
\]
\[
B(\boldsymbol{\tau})
:=
\left(
    C_{YZ}(\tau_1)-C_{YZ}(0),\,
    \ldots,\,
    C_{YZ}(\tau_a)-C_{YZ}(0)
\right)
\in \mathbb{R}^{d_Y \times a d_Z},
\]and $K = (\beta_{yy}, \beta_{yx}) \in \mathbb{R}^{d_Y \times a}$. The parameter 
matrix $K$ is uniquely determined iff $\mathrm{rank}(G(\tau)) = a$.

\subsection{Step 2: Trajectory spans \texorpdfstring{$\mathbb{R}^a$}{Ra}}

For the multi-input LTI system, we can determine the Kalman controllability of the pair $(A_c,b)$ via the rank condition of the controllability matrix $\mathcal{C}$, defined as:
\begin{equation}
    \mathcal{C} = \begin{pmatrix} b & A_c b & \cdots & A_c^{a-1} b 
    \end{pmatrix} \in \mathbb{R}^{a \times a d_Z}.
\end{equation}
This condition guarantees that every eigenmode of $A_c$ can be excited through some combination of input and delays. This equivalently means that the controllable subspace equals $\mathbb{R}^a$.

When $(A_c,b)$ is fully controllable and the diffusion of $Z$ is not degenerate, implying $\sigma_z\sigma_z^\top\succ0$, the trajectory could still fail to span $\mathbb{R}^a$, only if the initial condition is ``unluckily'' aligned with the system dynamics and control input. 

In standard control problems, the set of such initial conditions has measure zero when $x_0$ is randomly sampled, and thus usually this setting only obtains limited consideration. In our case however, $c_0$ is not chosen freely, but fixed by the full system dynamics via the stationary Lyapunov equation. Therefore, it remains to be shown that given the control input in form of the autocorrelation of process $Z$, the initial condition $c_0$ does not lie in such a subspace, in which the eigenmodes of $A_c$ are not fully visible in the trajectory. The following proof shows that given a non-degenerate instrument variation and controllability, the space covered by the $C(\tau)$ is guaranteed to stretch $\mathbb R^a$.

\xhdr{Lyapunov condition.} We first study the stationary covariance matrix $\Sigma$, which satisfies the Lyapunov equation
\begin{equation}
A \Sigma + \Sigma A^\top
=-\sigma\sigma^\top ,
\label{eq:Lyapunov equation}
\end{equation}
where
\begin{equation}
A =
\begin{pmatrix}
\beta_{yy} & \beta_{yx} & 0 & \beta_{yu}\\
\beta_{xy} & \beta_{xx} & \beta_{xz} & \beta_{xu}\\
0 & 0 & \beta_{zz} & 0\\
0 & 0 & 0 & \beta_{uu}
\end{pmatrix},
\qquad
\Sigma =
\begin{pmatrix}
\Sigma_{YY} & \Sigma_{YX} & \Sigma_{YZ} & \Sigma_{YU}\\
\Sigma_{YX}^\top & \Sigma_{XX} & \Sigma_{XZ} & \Sigma_{XU}\\
\Sigma_{YZ}^\top & \Sigma_{XZ}^\top & \Sigma_{ZZ} & 0 \\
\Sigma_{YU}^\top & \Sigma_{XU}^\top & 0 & \Sigma_{UU}
\end{pmatrix},
\end{equation}
and the block-diagonal diffusion matrix is given by
\begin{equation}
\sigma\sigma^\top
=
\operatorname{diag}
\!\left(
\sigma_y\sigma_y^\top,
\sigma_x\sigma_x^\top,
\sigma_z\sigma_z^\top,
\sigma_u\sigma_u^\top
\right).
\label{eq:diff-block}
\end{equation}
Evaluating sub-blocks of \cref{eq:Lyapunov equation} for $C(0) = c_0 = (\Sigma_{YZ},\Sigma_{XZ})$ and $C_{ZZ}(0) = \Sigma_{ZZ}$ results in:
\begin{equation}
\beta_{zz}\,\Sigma_{ZZ} + \Sigma_{ZZ}\,\beta_{zz}^\top
\;=\; -\,\sigma_z\sigma_z^\top.
\label{eq:laypunov_z}
\end{equation}
and
\begin{equation}
A_c\,c_0 \;+\; c_0\,\beta_{zz}^\top \;+\; b\,\Sigma_{ZZ}
\;=\; 0.
\label{eq:Lyapunov_cross}
\end{equation}

\xhdr{Span analysis.}
Next, we assume the existence of a direction $p$ that defines a subspace not covered by the trajectory, such that
\begin{equation}
    p^\top C(\tau) = 0 \quad \text{for all } \tau\geq 0 .
    \label{eq:cond_orthogonal_p}
\end{equation}

Now, we define $f(\tau) := p^\top C(\tau)$.
Since $C(\tau)$ is real-analytic in $\tau$, so is $f$ and the derivatives of $f$ exist everywhere on $[0,\infty)$. \Cref{eq:cond_orthogonal_p} implies that every derivative of $f$ is
zero, as the direction $p$ is never observed along the entire trajectory.
\[
f^{(k)}(0) = 0 \quad \text{for all } k \geq 0.
\]
We will show that this condition cannot be fulfilled under controllability by using the evaluations of $f^{(k)}(0)$ and the stationary Lyapunov conditions derived earlier, 
\cref{eq:laypunov_z} and \cref{eq:Lyapunov_cross}.

We now investigate the trajectories of the covariances $C$ and the instrument's lagged covariances jointly.
\begin{equation}
    \dot{C} = A_c C(\tau) + bC_{ZZ}(\tau), \qquad C(0) = c_0 , \qquad C_{ZZ}(\tau)=e^{\beta_{zz}\tau}\Sigma_{ZZ} .
\end{equation}

Starting with $f(0)=0$ we obtain the condition $p^\top c_0 = 0$. From the first derivative, we obtain: 
\begin{equation}
    p^\top \dot{C}(0)=0= p^\top A_c c_0 + p^\top b \Sigma_{ZZ} = p^\top (A_c c_0 + b \Sigma_{ZZ}) 
    \label{eq:first_dev}
\end{equation}
For the second derivative, we obtain after inserting the first derivative:
\begin{equation}
    p^\top \ddot{C}(0)=0 = p^\top A_c (A_c c_0 +b \Sigma_{ZZ}) +  p^\top b \beta_{zz} \Sigma_{ZZ}
    \label{eq:second_dev}
\end{equation}
Separately, we now multiply \cref{eq:Lyapunov_cross} with $p^\top A_c$: 
\begin{equation}
    p^\top(A_c^2 c_0 + A_c c_0 \beta_{zz}^T + A_c b \Sigma_{ZZ}) = 0
\end{equation}
and now use \cref{eq:first_dev} to obtain:
\begin{equation}
     p^\top (A_c^2 c_0 - b \Sigma_{ZZ} \beta_{zz}^T + A_c b \Sigma_{ZZ}) = 0
\end{equation}
Subtracting this equation from \cref{eq:second_dev}: 
\begin{equation}
    p^\top ( b \beta_{zz} \Sigma_{ZZ} + b \Sigma_{ZZ} \beta_{zz}^\top) = 0
\end{equation}
Now we can insert \cref{eq:laypunov_z}:
\begin{equation}
    p^\top b \sigma_z\sigma_z^T = 0
\end{equation}
Since we assume a non-degenerate diffusion for $Z$, we conclude: 
$p^\top b = 0$. 

\xhdr{Higher-order derivatives.}
We can now similarly cascade through the expressions for the higher derivatives. The general form for a higher derivative is given by: 
\begin{equation}
    C^{(m)}(0) = A_c^{m} c_0 + \sum_{j=0}^{m-1}A_c^{m-j-1} b \beta^j_{zz} \Sigma_{ZZ}
\end{equation}
We now proceed by induction and assume as induction hypothesis that 
$ p^\top A_c^i b = 0 $ for all $i \in \{0,\dots,k-1\} $.
For arbitrary $k+1, k+2$, we obtain first observe 
\begin{equation}
    p^\top C^{(k+2)}(0) = 0 = p^\top A_c^{k+2} c_0 + \sum_{j=0}^{k+1} p^\top A_c^{k+1-j} b \beta_{zz}^j \Sigma_{ZZ}
\end{equation}
We now use the induction hypothesis to kill all terms of $j$ for $j\geq2$, so that only the following remains:
\begin{equation}
    0 = p^\top A_c^{k+2} c_0 + p^\top A_c^{k+1} b \Sigma_{ZZ} + p^\top A_c^{k} b \beta_{zz} \Sigma_{ZZ}
    \label{eq:general_k+2}
\end{equation}
As before, we separately multiply \cref{eq:Lyapunov_cross} by $A_c^{k+1}$ and $p^\top$: 
\begin{equation}
    p^\top A_c^{k+2} c_0 + p^\top A_c^{k+1} c_0 \beta_{zz}^\top + p^\top A_c^{k+1} b \Sigma_{ZZ} = 0
\end{equation}
To obtain the value for $p^\top A_c^{k+1} c_0$, we multiply the general derivative at $m=k+1$ and use the inductive hypothesis for all terms with $j\geq 1$:
\begin{equation}
    p^\top C^{(k+1)}(0)=0 = p^\top A_c^{k+1} c_0 + p^\top A_c^{k} b \Sigma_{ZZ}    
\end{equation}
Substituting this yields:
\begin{equation}
    p^\top A_c^{k+2} c_0 - p^\top A_c^k b \Sigma_{ZZ} \beta_{zz}^T + p^\top A_c^{k+1} b \Sigma_{ZZ} = 0
    \label{eq:general_k+1}
\end{equation}

We now subtract \cref{eq:general_k+2} from \cref{eq:general_k+1} and obtain:
\begin{equation}
    p^\top A_c^{k}b(\beta_{zz} \Sigma_{ZZ} + \Sigma_{ZZ} \beta_{zz}^\top) = 0 
\end{equation}
Again using \cref{eq:laypunov_z}:
\begin{equation}
    p^\top A_c^k b \sigma_z \sigma_z^\top = 0
\end{equation}
which as before under a non-degenerate instrument diffusion implies $p^\top A_c^k b= 0$ and thus completes the induction.

By Cayley-Hamilton, we can express $A_c^a$ and higher order polynomials through the lower order polynomials, thus we do not obtain further conditions from higher derivatives and only need to consider all derivatives up to $a+1$. 

\begin{equation}
    p^\top \mathcal{C} = p^\top \begin{pmatrix} b & A_c b & \cdots & A_c^{a - 1} b 
    \end{pmatrix} = 0
\end{equation}
with $\mathcal{C} \in \mathbb{R}^{a \times ad_Z}$.

If $\mathcal{C}$ has full row rank, it has a trivial left nullspace, and no such direction $p^\top$ exists.

Consequently, the full rank of $\mathcal{C}$ guarantees that the trajectory of $C(\tau)$ spans $\mathbb R^a$, and the degeneracy of a badly chosen initial condition cannot occur.

\subsection{Step 3: Sampling points \texorpdfstring{$\boldsymbol\tau$}{tau}}
If the trajectory of $C(s)$ spans $\mathbb R^a$, the same is true for the integrated covariance curve, as any direction orthogonal to its integrals would also need to be orthogonal to the covariance curve itself. Hence, tuples of $a$ timepoints, where $\mathrm{rank}(G)=a$ exist. Degenerate choices of sampling points can still occur. Intuitively, this happens, when the sampled times repeatedly hit an eigenfrequency of the system, such that the columns of $G$ are linearly dependent. However, such sampling points lie on a lower-dimensional subset of the sampling space. Thus, as the timepoints are drawn from a continuous distribution over positive sampling times, this set has Lebesgue measure zero. 

Consequently $a$ randomly drawn lags
almost surely give $\mathrm{rank}(G(\boldsymbol\tau)) = a$ and identify $K$.

\xhdr{Note: Sufficiency of less timepoints.} Each instrument process $Z$ contributes one moment condition at every sampled timepoint $\tau$, giving a total of $ad_z$ moment conditions for $a$ sampled timepoints. Compared to the $a$ parameters that are identified, this generally already puts the linear equation system \cref{eq:lin_sys_prop} in the overidentified setting, if $d_z>1$. The reason for this is that the correlation with different instrument processes might excite different and partially overlapping eigenmodes of matrix $A_c$. Under the controllability criterion, each instrument excites at most $a$ eigenmodes, and all instruments jointly excite exact $a$ eigenmodes. Thus, given controllability, the number of timepoints required for the just identified setting lies between 
$a/d_Z$ and $a$. Sampling from all instrument processes at $a$ times thus gives a conservative identification strategy.

\section{Estimator Consistency in the Linear Case}
\label{app:estimator_discussion}

We present two estimators based on either the integral weak form formulation of the covariance ODE or the differential formulation. We show consistency for the integral form of the estimator.

\xhdr{Integral linear IV estimator.} The fixed number of lags $0<\tau_1<\cdots<\tau_L$, with $L>a$, is drawn from a continuous distribution on $(0,\tau_{max}]$, as defined in \cref{prop:controllability}. Further, denote

\begin{equation}
    V_t:=\begin{pmatrix}Y_t\\X_t\end{pmatrix},
    \qquad
    C(s):=\mathbb E[V_{t+s}Z_t^\top]
    =\begin{pmatrix}C_{YZ}(s)\\C_{XZ}(s)\end{pmatrix}.
\end{equation}

We define

\begin{equation}
    B_\ell:=C_{YZ}(\tau_\ell)-C_{YZ}(0),
    \qquad
    G_\ell:=\int_0^{\tau_\ell}C(s)\,ds,
    \label{eq:draft-integral-blocks}
\end{equation}
and stack the blocks as $B=(B_1,\ldots,B_L)$ and
$G=(G_1,\ldots,G_L)$. 
Integrating the covariance ODE gives
\begin{equation}
    B=KG,\qquad \text{with } K=(\beta_{yy},\beta_{yx}).
    \label{eq:draft-integral-moments}
\end{equation}

We assume $(Y,X,Z)$ are observed at times $t_1=0,t_2,\ldots,t_{n}$,
with positive, i.i.d. gaps $\Delta_i=t_{i+1}-t_i$ having known density $q$, which is bounded,
positive on $(0,T]$, continuous and independent of the process. We may then define k-timestep gaps as $\Delta_{i,k}:=t_{i+k}-t_i$ and, considering the next $m$ samples, compute the expected number of successive samples at each time gap $s$ as $\rho(s):=\sum_{k\ge1}^mq^{*k}(s)$. We choose $mT>\tau_{max}$, such that our sampling interval covers the maximum lag to be evaluated.
Each integrated block is then estimated by
\begin{equation}
\widehat G_{\ell,n}
:=
\frac1{n-m}\sum_{i=1}^{n-m}\sum_{k=1}^{m}
\frac{\mathbbm{1}\{\Delta_{i,k}\le\tau_\ell\}}
     {\rho(\Delta_{i,k})}
V_{t_{i+k}}Z_{t_i}^{\top}.
\label{eq:draft-g-estimator}
\end{equation}

Following~\cite{Walk2010}, we define a kernel $\mathcal K$ and bandwidth $h_n>0$, with
$\mathcal K_h(v):=h^{-1}\mathcal K(v/h)$. The bandwidths $h_n>0$ are chosen as $h_n = c_mn^{-\gamma}$, with constants $c_m>0$ and $0<\gamma<1$. We use this kernel to define a Nadaraya-Watson estimator to nonparametrically regress the covariance function at the chosen lags $\tau_\ell$. The covariance function at the lag does not admit an estimator in the form of a simple sample average, as we almost surely do not observe a sample at exactly $\tau_\ell$, thus, we use this nonparametric first stage. The regression function takes the form of $g(\Delta) = \mathbb E(Y_{t+\Delta}\,Z_t^T|\Delta)$. Further, we define $J_\ell:=\{k\in\{1,\ldots,m\}:q^{*k}(\tau_\ell)>0\}$ which selects the subsequent point indices, whose lag distributions are positive at the target lag $\tau_\ell$. We then estimate the covariance function at lag zero and each sampled lag $\tau_\ell$ as follows:
\begin{equation}
\widehat C_{YZ,n}(\tau_\ell)
:=
\frac{
\sum_{i=1}^{n-m}\sum_{k\in J_\ell}
\mathcal K_{h_n}(\Delta_{i,k}-\tau_\ell)
Y_{t_{i+k}}Z_{t_i}^{\top}
}{
\sum_{i=1}^{n-m}\sum_{k\in J_\ell}
\mathcal K_{h_n}(\Delta_{i,k}-\tau_\ell)
},
    \qquad
    \widehat C_{YZ,n}(0):=\frac1n\sum_{i=1}^nY_{t_i}Z_{t_i}^\top.
    \label{eq:draft-endpoint-estimator}
\end{equation}
Further, we denote
\begin{equation}
    \widehat B_{\ell,n}
        :=\widehat C_{YZ,n}(\tau_\ell)-\widehat C_{YZ,n}(0),
    \qquad
    \widehat B_n:=(\widehat B_{1,n},\ldots,\widehat B_{L,n}),
\end{equation}
and $\widehat G_n:=(\widehat G_{1,n},\ldots,\widehat G_{L,n})$.
We fit the sample moments by least squares. The analytic minimizer of \cref{eq:draft-integral-moments} can be obtained as
\begin{equation}
    \widehat K_n
    :=\widehat B_n\widehat G_n^\top
        (\widehat G_n\widehat G_n^\top)^{-1},
    \label{eq:draft-integral-ls-estimator}
\end{equation}

which gives the final parameter estimates for the integral linear estimator.

\consistencyint*

\begin{proof}
We start by showing the consistency of the integrated blocks.
Stationarity and ergodicity, together with finite second moments of the process imply that $\widehat C_{YZ,n}(0)\overset{p}{\longrightarrow}C_{YZ}(0)$.

To show consistent estimation of $\widehat{G}$, we first notice that $\mathbb E[V_{t_{i+k}}Z_{t_i}^{\top}\mid\Delta_{i,k}=s]=C(s)$ and write
\begin{equation}
\begin{aligned}
G_\ell
&= \int_0^{\tau_\ell} C(s)\,ds \\
&= \sum_{k=1}^{m}\int_0^{\tau_\ell}
   \frac{\mathbb E[V_{t+s}Z_t^\top]}{\rho(s)}
   q^{*k}(s)\,ds \\
&= \sum_{k=1}^{m}\mathbb E\left[
   \frac{\mathbbm{1}\{\Delta_{i,k}\le\tau_\ell\}}
        {\rho(\Delta_{i,k})}
   \mathbb E\left[
      V_{t_{i+k}}Z_{t_i}^\top\mid\Delta_{i,k}
   \right]
   \right] \\
&= \mathbb E\left[
   \sum_{k=1}^{m}
   \frac{\mathbbm{1}\{\Delta_{i,k}\le\tau_\ell\}}
        {\rho(\Delta_{i,k})}
   V_{t_{i+k}}Z_{t_i}^\top
   \right]
 = \mathbb E[W_{i,\ell}].
\end{aligned}
\end{equation}
We then estimate the $\widehat{G}_{\ell,n}$ via the sample average as described in \cref{eq:draft-g-estimator}.
Using the independence of the observation times from the process implies that the process $W_{i,\ell}$ is stationary and ergodic. Further, we can show integrability by
using Cauchy Schwarz and stationarity
\begin{equation}
    \mathbb E|W_{i,\ell}| \le \int_0^{\tau_{\ell}}\mathbb{E} |V_{t+s}Z_t^T|ds \leq \tau_\ell \sqrt{\mathbb{E}|V_t|^2\mathbb{E}|Z_0|^2}< \infty,
\end{equation}       
where the first inequality comes from using the triangle inequality over the sum up to $m$ timepoints. This thus allows to invoke Birkhoff's ergodicity theorem~\citep{durrett2019probability} and yields $\widehat G_{\ell,n}\overset{p}{\longrightarrow}G_\ell$. 

Following~\cite[Thm. 3b]{Walk2010}, we show the consistency of the kernel regression of $C_{YZ}(\tau_\ell)$ with either a uniform or Gaussian kernel. First notice that the first equation in \cref{eq:draft-endpoint-estimator} can be written as a weighed sum of Nadaraya-Watson estimators for each $k\in J_\ell$.
\begin{equation}
\widehat C_{YZ,n}(\tau_\ell)
=
\sum_{k\in J_\ell}
w_{k,\ell}\widehat C^{(k)}_{YZ,n}(\tau_\ell),
\qquad
w_{k,\ell}
:=
\frac{
\sum_{i=1}^{n-m}
\mathcal K_{h_n}(\Delta_{i,k}-\tau_\ell)
}{
\sum_{j\in J_\ell}\sum_{i=1}^{n-m}
\mathcal K_{h_n}(\Delta_{i,j}-\tau_\ell)
},
\end{equation}
with 
\begin{equation}
\widehat C^{(k)}_{YZ,n}(\tau_\ell)
:=
\frac{
\displaystyle\sum_{i=1}^{n-m}
\mathcal K_{h_n}(\Delta_{i,k}-\tau_\ell)
Y_{t_{i+k}}Z_{t_i}^{\top}
}{
\displaystyle\sum_{i=1}^{n-m}
\mathcal K_{h_n}(\Delta_{i,k}-\tau_\ell)
}.
\label{eq:fixed-k-endpoint-estimator}
\end{equation}
For each $k$ the sequence of samples $(\Delta_{i,k},Y_{t_{i+k}}Z_{t_i}^\top)$ is geometrically mixing, as seen from the OU transition matrix, which takes the form of a matrix exponential. Geometric mixing is stronger than the polynomial mixing required by by \citet{Walk2010}. Thus, we may choose $h_n=c_m(n-m)^{-\gamma}$ with any $0<\gamma<1$ and $c_m>0$. Consequently, each of the estimators $\widehat C^{(k)}_{YZ,n}(\tau_\ell)$ fulfills the conditions presented in \citet{Walk2010} and is consistent.
As the weights are nonnegative and sum up to one, the average of these estimators and thus the final estimate for the endpoint covariance are consistent as well. 
This yields 
$\widehat C_{YZ,n}(\tau_\ell)\overset{p}{\longrightarrow}C_{YZ}(\tau_\ell)$,
as well as $\widehat B_{\ell,n}\overset{p}{\longrightarrow}B_\ell$ through the continuous mapping theorem, and for the stacked matrices
$\widehat G_n\overset{p}{\longrightarrow}G$ and
$\widehat B_n\overset{p}{\longrightarrow}B$.

%\textit{Consistency of the second stage given the first stage}

If the identification condition defined in~\cref{eq:controllability_rank} holds, matrix $G$ has full rank. Hence, $GG^T$ is almost surely invertible. We use the least-squares solution for the final estimate of the parameter matrix. Following the continuous-mapping theorem, this function is continuous around an open set of the true matrix $GG^T$. Thus, given the consistent estimation of $\widehat{B}$ and $\widehat{G}$, the estimate of $K$ is consistent.
\begin{equation}
    \widehat K_n
    \overset{p}{\longrightarrow}BG^\top(GG^\top)^{-1}=K.
\end{equation}
\end{proof}

\xhdr{On a simulation grid.}
The nonparametric first stage using the kernel-based estimation of $C_{YZ}(\tau_\ell)$ can be replaced by a simple sample average in case of working on a simulation grid with random dropouts defined by a simulation interval $\delta_{sim}$ and a uniformly random dropout percentage per point $p_{drop}$, as in such a setup each $\tau_\ell$ has a non-zero chance of being sampled. In this case, we can fix the kernel bandwidth $h$ to $0<h<\delta_{sim}$, which retrieves the standard sample average for the covariance estimation at the given lag $\tau_\ell$. We note that this discrete version does not keep the previously presented consistency result due to numeric approximation errors.

\xhdr{Differential linear IV estimator.} As a second estimator, we propose the differential estimator.
We define the differential IV estimator by binning the covariances across lags to estimate the covariance curves on a fine grid in the first stage. In the second stage we then differentiate the curves at the lags $0<\tau_1<\cdots<\tau_L$, which are sampled as previously described, and then solve an equation system based on the differential form of the covariance ODE via least squares.
The preceding consistency argument for the integral estimator does not extend directly to this differential form. Establishing consistency would require a separate analysis of the derivative estimator given the
specific smoothing and sampling schemes.

\section{Influence of a non-diagonal diffusion matrix.}
\label{app:non_diag_diffusion_matrix}
In this section, we discuss the effects of a non-diagonal diffusion matrix for the linear case specifically.
As mentioned in~\cref{app:setup}, the moment conditions themselves remain valid assuming a non-diagonal diffusion matrix, as long as exogeneity of the instrument is fulfilled. Specifically, in case of an OU process this requires $\bigl(\sigma(S_t)\sigma(S_t)^\top\bigr)_{zu} = 0$. 

However, we note that the identification argument presented in \cref{prop:controllability} might not hold in case of diffusion mixing of $Y$ and $Z$. In this case, even under the required controllability condition, the initial condition derived via the Lyapunov equation, as done in \cref{proof:prop_2}, can be unluckily chosen such that the resulting covariance curves do not cover the required state space of $\mathbb R^a$. Consequently, in this case the moment equations do not necessarily identify the drift function anymore.

Nevertheless, we empirically investigated the effect of two special cases of non-diagonal diffusion matrices in OU-processes on our proposed estimation, as shown in \cref{tab:diffusion_mixing_mse}.

\begin{table}[th]
    \centering
    \caption{Total MSE of the IV-SDE estimator under different diffusion configurations and mixing strengths.}
    \label{tab:diffusion_mixing_mse}
    \begin{tabular}{lc}
        \toprule
        Configuration & IV-SDE MSE (total) \\
        \midrule
        \(x-y,\ k=0.0\)             & 0.0036 \\
        \(x-y,\ k=0.3\)             & 0.0068 \\
        \(x-y,\ k=0.5\)             & 0.0151 \\
        \(x-y,\ k=0.9\)             & 0.0162 \\
        \(z-x,\ k=0.3\)             & 0.0031 \\
        \(z-x,\ k=0.5\)             & 0.0009 \\
        \(z-x,\ k=0.9\)             & 0.0011 \\
        Violation (dense, \(k=0.6\)) & 0.2306 \\
        \bottomrule
    \end{tabular}
\end{table}

In the first case, we mix the diffusions between $X$ and $Y$ with increasing strength up to the level of the diagonal entry, thus $\sigma_{xy,k}=k\sigma_x$ and $\sigma_{yx,k}=k\sigma_y$. The results show that with increasing mixing strength the MSE of the estimation increases. Stronger mixing increases the variance of the outcome and thereby increases the variance of the moment conditions, resulting in lower estimation efficiency.

In a second experiment, we mix the diffusions between $Z$ and $X$. Interestingly, moderate mixing may decrease the MSE of the estimation. The increased coupling between process $X$ and $Z$ increases the effective instrument strength and thus allows for a better effect estimation. Beyond a certain mixing level, however, the increase in the overall estimation variance dominates this gain in relevance, causing the estimation error to increase again.
Further, the experiments confirm that a dense covariance matrix violates the assumptions, resulting in a higher estimation error.

\section{The signature kernel}
\label{app:sig_kernel}
The signature kernel~\citep{lee2023signaturekernel,manten2025signaturekernelconditionalindependence} relies on the signature features $\mathbf{S}(X) \in \mathcal T(\mathbb{R}^d)$ of continuous bounded variation paths $X \in C([0,T], \mathbb{R}^{d})$, where $\mathcal T$ describes the tensor algebra over $\mathbb{R}^d$. These features can be understood as a sequence of n-fold iterated integrals of the path $X$ against itself, which together give a universal feature map of continuous paths. The signature kernel $k_z: (C([0,T], \mathbb{R}^{d}) \times C([0,T], \mathbb{R}^{d}) \to \mathbb R$ between paths $X, Y$ is the tensor-algebra inner product of their signatures, $k_{z}(X, Y) = \langle \mathbf{S}(X), \mathbf{S}(Y) \rangle$.
The kernel can be efficiently evaluated without explicitly computing the signature features by using the kernel trick, which then requires solving an associated hyperbolic PDE~\citep{Salvi_2021}.

\section{Experimental Details}
\label{app:exp_details}

The linear experiments were performed on CPU. The nonlinear 
experiments were run on a single A100 GPU in under two hours for all 
evaluations.

\subsection{Linear (OU) Experiments}
The linear experiments were all simulated with the following SDE system:
\begin{equation}
    d \mathbf{S}_t = 
    \underbrace{
    \begin{pmatrix} 
        \beta_{yy} & \beta_{yx} & 0 & \beta_{yu} \\ 
        \beta_{xy} & \beta_{xx} & \beta_{xz} & \beta_{xu} \\ 
        0 & 0 & \beta_{zz} & 0 \\ 
        0 & 0 & 0 & \beta_{uu} 
    \end{pmatrix}
    }_{A}
    \mathbf{S}_t \, dt 
    + 
    \underbrace{
    \begin{pmatrix} 
        \sigma_y & 0 & 0 & 0 \\ 
        0 & \sigma_x & 0 & 0 \\ 
        0 & 0 & \sigma_z & 0 \\ 
        0 & 0 & 0 & \sigma_u 
    \end{pmatrix}
    }_{\boldsymbol{\sigma}}
    \begin{pmatrix} dW^y_t \\ dW^x_t \\ dW^z_t \\ dW^u_t \end{pmatrix}.
    % \label{eq:IV_OU_structure}
\end{equation}
where $\beta_{ab} \in \mathbb{R}^{d_a \times d_b}$ for $a, b \in \{y, x, z, u\}$. 
Unless stated otherwise, we use $d_y = d_x = d_z = d_u = 1$.

\xhdr{Parameter sampling.} All non-zero drift coefficients are sampled 
uniformly from $[-1, -0.2]\cup[0.2,1]$, ensuring existing edges of the sparse connectivity graph are bounded away from zero. 
Diffusion coefficients are sampled uniformly from $[0.2, 0.5]$. All sampled 
systems are filtered for stability (negative real parts of all eigenvalues 
of $A$) and controllability (full rank of the controllability matrix 
$\mathcal{C}$ defined in \cref{eq:controllability_rank}).

\xhdr{Simulation.} For the left plot in \Cref{fig:combined_all}, trajectories are simulated by sampling from the analytic transition densities at irregularly sampled observation times. The irregular observation density $q$ is set to be uniform on an interval of $(0,0.4]$. 
We evaluate each method on 8000 sample paths of length $T_{max} = 20\,\mathrm{s}$, initialized from the analytically determined stationary distribution. 
For the plots in \Cref{fig:combined_all} in the middle and right, we sampled the trajectories on a simulation grid of $0.01s$, kept a random small subset of points and used the IV-estimator on a simulation grid, as described in \cref{app:estimator_discussion}.

\xhdr{Experiment-specific settings.} In the confounding strength 
experiment, we simultaneously vary $\beta_{xu}$ and $\beta_{yu}$ over the 
range $[0.1, 1.0]$, sampling 200 random systems across all confounding values. 
For the covariate dimension experiment, we sample 30 systems in total. 
For the instrument strength experiment, we sample 80 systems in total, varying 
$\beta_{xz}$ over $[0.1, 0.8]$.

\xhdr{Baselines.} We note that the only available baselines, besides the naive MLE estimator, for causal IV methods in time are presented in \cite{mogensen2023} and \cite{thams2024}, both of which are designed for discrete time methods. As both were not designed to handle irregularly spaced timepoints in continuous-time systems, we cannot directly compare them fairly on the same data. Thus, besides applying the $\mathrm{VAR}(1)$ estimator naively on irregular samples, we also sample the same number of regularly spaced timepoints from the simulated trajectories, and apply the discrete-time estimators on these.

To construct the MLE estimator, we assume the full system to be Markov with respect to just the observed states $(Y,X,Z)$ and use the analytically available and exact Gaussian transition densities to estimate the drift parameters.

We applied the discrete $\mathrm{VAR}(1)$ estimator on regularly sampled timepoints with spacing of $0.2s$ and match the number of continuous-time timepoints. This yields an estimate for the discrete-time transition matrix. We estimate the continuous-time target parameters from the transition matrix by applying the matrix logarithm and dividing by the sampling interval. We also naively apply the discrete time estimator on the irregular timepoints.

We further compare to the discrete-time IV method presented~\cite{mogensen2023} which estimates the \textit{normalized parameters} using integrated covariance matrices. Here, the \textit{normalized parameters} correspond to the ratio $\beta_{yx}/\beta_{yy}$. The method does not identify the target parameters themselves, thus we need to compare our results to their ratio estimate.  We also apply this method on the regularly spaced timepoints.

\xhdr{Extended results discussion.} \Cref{tab:coefficient-error} shows the mean coefficient error across all 200 simulated systems, with 20 sampled systems per confounding strength and irregular spacing. The results clearly demonstrate smaller estimation errors of the IVSDE estimators under stronger confounding. Both IV estimators, the differential IVSDE estimatior (IVSDE-diff), which bins the data across 0.05s bins, and the continuous-time integral IVSDE estimator (IVSDE-int) presented in \cref{app:estimator_discussion} perform similarly well on the data. In case of minimal or no confounding, the MLE yields slightly lower estimation errors, as it uses the full transition densities and is thus more efficient. The naively applied $\mathrm{VAR}(1)$ method fails on irregularly spaced data.

The results for the direct ratio estimation presented in \cref{tab:ratio-error} suffer from the same discrete sampling bias as the discrete $\mathrm{VAR}(1)$ method. As a consequence, using the proposed IVSDE method to first individually estimate the parameters and then construct ratio estimates still outperforms this direct ratio estimation, despite being a more complex two-step estimation.

\begin{table*}[th]
    \centering
    \small
    \setlength{\tabcolsep}{3pt}
    \caption{Mean per-system parameter RMSE over 20 systems. (reg.) denotes that regularly sampled timepoints were used.}
    \label{tab:coefficient-error}
    \begin{tabular}{lccccc}
        \toprule
        Confounding strength & MLE full & VAR-1 & VAR-1 (reg.) & IVSDE-diff & IVSDE-int \\
        \midrule
        0.1 & \textbf{0.0069} & 0.6590 & 0.0516 & 0.0104 & 0.0123 \\
        0.2 & 0.0379 & 1.2766 & 0.0625 & \textbf{0.0148} & 0.0192 \\
        0.3 & 0.0691 & 1.5407 & 0.0576 & \textbf{0.0182} & 0.0202 \\
        0.4 & 0.1244 & 1.1334 & 0.0464 & 0.0244 & \textbf{0.0176} \\
        0.5 & 0.1146 & 0.8761 & 0.0431 & \textbf{0.0097} & 0.0137 \\
        0.6 & 0.1746 & 0.8853 & 0.0613 & 0.0214 & \textbf{0.0184} \\
        0.7 & 0.2065 & 0.9870 & 0.0456 & \textbf{0.0112} & 0.0136 \\
        0.8 & 0.2112 & 0.6470 & 0.0506 & 0.0225 & \textbf{0.0126} \\
        0.9 & 0.2397 & 1.0516 & 0.0675 & 0.0255 & \textbf{0.0185} \\
        1.0 & 0.2359 & 0.9859 & 0.0616 & 0.0309 & \textbf{0.0290} \\
        \bottomrule
    \end{tabular}
\end{table*}

  \begin{table*}[th]
      \centering
      \caption{Ratio RMSE. Ratio RMSE across varying confounding strength. Each value represents the mean over 20 sampled systems.}
      \label{tab:ratio-error}
      \setlength{\tabcolsep}{3pt}
      \begin{tabular}{lcccccc}
          \toprule
          Confounding strength & MLE full & VAR-1 & VAR-1 (reg.) & IVSDE-diff & IVSDE-int & Direct ratio (reg.) \\
          \midrule
          0.1 & 0.0472 & 0.8777 & 0.3202 & \textbf{0.0166} & 0.0174 & 0.1005 \\
          0.2 & 0.1711 & 1.9936 & 0.1894 & \textbf{0.0504} & 0.0526 & 0.0613 \\
          0.3 & 1.2192 & 2.2831 & 0.1939 & 0.0772 & \textbf{0.0649} & 0.3011 \\
          0.4 & 0.3261 & 2.1911 & 0.2962 & \textbf{0.0218} & 0.0233 & 0.1542 \\
          0.5 & 0.5827 & 1.6280 & 0.2167 & \textbf{0.0338} & 0.0348 & 0.1127 \\
          0.6 & 1.1755 & 1.2828 & 0.2407 & \textbf{0.0344} & 0.0442 & 0.1520 \\
          0.7 & 2.6015 & 3.6445 & 0.2048 & 0.0305 & \textbf{0.0261} & 0.2258 \\
          0.8 & 15.2417 & 3.0491 & 0.5627 & 0.0615 & \textbf{0.0566} & 0.2535 \\
          0.9 & 84.2825 & 2.4190 & 0.3090 & 0.0916 & \textbf{0.0794} & 0.1910 \\
          1.0 & 1.7591 & 1.4126 & 0.2655 & 0.0462 & \textbf{0.0372} & 0.2251 \\
          \bottomrule
      \end{tabular}
  \end{table*}

\subsection{Kernel-based estimation of a basis-function expansion}

\xhdr{Simulation.}
We simulate paths from the following nonlinear SDE system, where all processes are one-dimensional except the instrument process $Z_t$, which is three-dimensional ($d_z = 3$):
\begin{align}
    dY_t &= \left(\alpha Y_t + \beta X_t + \sum_{i=1}^p w_i e^{-(X_t - c_i)^2/\sigma_{\mathrm{rbf}}^2} - \sum_{i=1}^p w_i e^{-(X_t + c_i)^2/\sigma_{\mathrm{rbf}}^2} + \beta_{yu} U_t\right)dt + \sigma_y \, dW_t^y \\
    dX_t &= \left(\beta_{xz} Z_t + \beta_{xx} X_t + \beta_{xu} U_t\right)dt + \sigma_x \, dW_t^x \\
    dZ_t &= \beta_{zz} Z_t \, dt + \sigma_z \, dW_t^z \\
    dU_t &= -a_u U_t \left(U_t^2 - 0.5^2\right)dt + \sigma_u \, dW_t^u
\end{align}
where the confounding process $U_t$ follows a nonlinear mean-reverting dynamics, resulting in a non-Gaussian distribution of $X$. The drift of $Y$ is parameterized using two linear expressions in $X$ and $Y$, and pairs of radial basis functions symmetric about the origin, which preserves the zero mean of process $Y$.
As described in the main text, we assume access to densely sampled observations and therefore use the time series directly on the simulation grid. In settings with coarser sampling, simulation-based estimation methods would be required to reliably approximate the integral terms in the residual $R_{t,t+h;\theta}$. 

\xhdr{Basis function parametrization.}
We parametrize the target drift $f_{y;\theta}$ using a basis function expansion consisting of $p = 4$ radial basis functions (RBFs) with centers $\{c_i\}_{i=1}^4$ placed uniformly over $[0.2, 2]$, covering approximately $96\%$ of the marginal distribution of $X_t$ under stationarity, together with two linear terms in $X_t$ and $Y_t$. The bandwidth of each RBF is set to the ground truth $\sigma_{\mathrm{rbf}} = 0.35$. The resulting parametrization is:
\begin{equation}
    f_{y;\theta}(Y_t, X_t) = \alpha Y_t + \beta X_t + \sum_{i=1}^4 w_i e^{-(X_t - c_i)^2/\sigma_{\mathrm{rbf}}^2} - \sum_{i=1}^4 w_i e^{-(X_t + c_i)^2/\sigma_{\mathrm{rbf}}^2}
\end{equation}
where $\theta = (\alpha, \beta, w_1, \ldots, w_4) \in \mathbb{R}^6$ are the parameters to be estimated.

\xhdr{Closed-form estimation.}
Given the parametrization $f_{y;\theta}(Y_t, X_t) = B_t^\top \theta$, which is linear in parameters, where $B_t \in \mathbb{R}^{p}$ collects the basis function evaluations at time $t$, the residual $R_{t,t+h;\theta}$ is linear in $\theta$:
\begin{equation}
    R_{t,t+h;\theta} = Y_{t+h} - Y_t - \int_t^{t+h} B_s^\top \theta \, ds = Y_{t+h} - Y_t - \bar{B}^\top_{t,h} \theta
\end{equation}
where $\bar{B}_{t,h} = \int_t^{t+h} B_s \, ds \in \mathbb{R}^p$ is the time-integrated basis function vector over the horizon $h$, approximated via numerical quadrature on the dense simulation grid, and $\mathbf{e}$ describes the difference $Y_{t+h}-Y_t$. As a result, the V-statistic risk $\widehat{\mathcal{R}}_z(\theta)$ in \cref{eq:finalobjective} reduces to a quadratic objective in $\theta$, and the minimization problem admits the closed-form kernel ridge regression solution:
\begin{equation}
    \hat{\theta} = \left(\bar{\mathbf{B}}^\top K_z \bar{\mathbf{B}} + \lambda I\right)^{-1} \bar{\mathbf{B}}^\top K_z \mathbf{e}
    \label{eq:nonlin_closed_form}
\end{equation}
where $\bar{\mathbf{B}} \in \mathbb{R}^{n \times p}$ stacks the integrated basis vectors $\bar{B}_{t_i, h}$ across all $n$ sampled path sections, $K_z \in \mathbb{R}^{n \times n}$ is the signature kernel Gram matrix evaluated on the instrument paths $\{Z^i_{[0,t]}\}_{i=1}^n$, $\mathbf{e} \in \mathbb{R}^n$ collects the observed outcome increments $Y^i_{t+h} - Y^i_t$, and $\lambda > 0$ is the regularization parameter. This allows the full estimation to be performed in a single linear solve, without iterative optimization.

\xhdr{Sampling scheme.}
We sample $5000$ path segments, each consisting of the forward prediction horizon $h$ and the lookback time $\ell$ over $Z$, which is used as the instrument path. We use the following hyperparameters: prediction horizon $h = 0.3\,\text{s}$, stride between consecutive sections $= 2\,\text{s}$, and lookback window for the instrument path $\ell= 1.5\,\text{s}$ and $\lambda=0.001$.

\subsection{Nonlinear drift function with a neural network}
Furthermore, we show that the kernel-based estimation scheme also allows for a neural parameterization of the drift function. 
We keep the same setup for the instruments and 
covariates as described in the previous section. The drift of $Y$ now depends on $X$ 
via a sigmoid nonlinearity $S$, resulting in the following:
\begin{equation}
    dY_t = (\alpha Y_t + S(kX_t)+U_t)\,dt + \sigma_y\,dW_t^y
\end{equation}

where we choose k=4. We parameterize $f_\theta$ with a small MLP with two inputs $(X_t, Y_t)$, two hidden layers of dimension 16, and tanh activation functions.

We sample $n = 10000$ path segments and 
optimize the empirical risk in~\cref{eq:finalobjective} with the following hyperparameters using the Adam optimizer: 
integration horizon $h = 0.3\,\text{s}$, 
lookback $\ell = 1.5\,\text{s}$, weight decay $\lambda = 1 \times 10^{-4}$, learning 
rate $\eta = 10^{-3}$, and $N = 1000$ gradient steps.

\subsection{Diabetes dataset.}
\xhdr{Setup.} We setup a semi-synthetic experiment to demonstrate the applicability of our proposed method on problems of real-world complexity and relevance. We motivate our setting from pharmacokinetics / pharmacodynamics models of type 2 diabetes. These models describe the dynamics and interactions of insulin and glucose concentrations in the blood. In type 2 diabetes specifically, one specific mechanism of the disease is found in the insulin resistance build up by patients, as muscle and fat cells take up glucose less efficiently. Consequently, identifying the insulin sensitivity in such setting allows to judge a patient's personal disease progression and guide future insulin dosing.
Both insulin production as well as blood glucose concentration can be confounded by unobserved gut glucose. Here, the insulin production is stimulated by incretin hormones, such as GLP-1, which are secreted by enteroendocrine cells in the gut, while intestinal glucose absorption provides a direct additive contribution to the blood glucose concentration.

The setup thus fulfills the requirements of our setting, with scheduled external insulin dosages as instruments, the insulin concentration as the covariate process, glucose concentration as the outcome process, and the insulin sensitivity as the estimand. It illustrates our method in a system combining nonlinear interactions, feedback, and an unobserved common driver.

We follow the structural setup of the model presented in \cite{Naik_2026, Subramanian_2024}, select the parameters for the simulation setting such that we obtain a stable SDE and set the instrument process to be a two-dimensional OU process, resulting in the following SDE system 

\begin{equation}
\begin{aligned}
dZ_{j,t}
&=
m_{Z,j}\bigl(\mu_Z-Z_{j,t}\bigr)\,dt
+\sigma_{Z,j}\,dW_{t}^{Z_j},
\qquad j=1,\ldots,k,
\\[4pt]
dX_t
&=
\left(
\sum_{j=1}^{k} a_{Z,j} Z_{j,t}
+c_{ux}U_t
+c_{yx}Y_t
-a_x X_t
\right)\,dt
+\sigma_x\,dW_{t}^x,
\\[4pt]
dY_t
&=
\left(
c_{yu}U_t
-a_y Y_t
-\gamma_{xy}X_tY_t
\right)\,dt
+\sigma_y\,dW_{t}^y,
\\[4pt]
dG_t
&=
\left(
\mu_g-a_g G_t
\right)\,dt
+\sigma_g\,dW_{t}^g,
\\[4pt]
dU_t
&=
\left(
g_{gu}G_t-a_uU_t
\right)\,dt
+\sigma_u\,dW_{t}^u.
\end{aligned}
\end{equation}

\begin{table}[htbp]
\centering
\caption{Parameter values for the two-instrument diabetes dataset configuration.}
\label{tab:diabetes_parameters}
\begin{tabular}{lll}
\hline
Parameter & Description & Value \\
\hline
$k$ & Number of instruments & $2$ \\
$\mu_Z$ & Common instrument equilibrium mean & $0$ \\
$(m_{Z,1},m_{Z,2})$ & Instrument mean-reversion rates & $(0.4,\,0.8)$ \\
$(a_{Z,1},a_{Z,2})$ & Instrument effects on $X$ & $(0.35,\,0.35)$ \\
$(\sigma_{Z,1},\sigma_{Z,2})$ & Instrument noise amplitudes & $(0.5,\,0.5)$ \\
$c_{ux}$ & Effect of $U$ on $X$ & $0.60$ \\
$c_{yx}$ & Effect of $Y$ on $X$ & $0.10$ \\
$a_x$ & Decay rate of $X$ & $0.30$ \\
$\sigma_x$ & Noise amplitude of $X$ & $0.02$ \\
$c_{yu}$ & Effect of $U$ on $Y$ & $2.10$ \\
$\mathbf{a_y}$ & \textbf{Decay rate of $Y$} & $0.10$ \\
$\boldsymbol{\gamma_{xy}}$ & \textbf{Insulin sensitivity} & $0.15$ \\
$\sigma_y$ & Noise amplitude of $Y$ & $0.02$ \\
$\mu_g$ & Constant inflow into $G$ & $0.05$ \\
$a_g$ & Decay rate of $G$ & $0.05$ \\
$\sigma_g$ & Noise amplitude of $G$ & $0.005$ \\
$g_{gu}$ & Effect of $G$ on $U$ & $0.05$ \\
$a_u$ & Decay rate of $U$ & $0.05$ \\
$\sigma_u$ & Noise amplitude of $U$ & $0.02$ \\
\hline
\end{tabular}
\end{table}
  \begin{table}[th]
      \centering
      \caption{Hyperparameter sensitivity in the diabetes model.
      in terms of RMSE. The standard setting across all runs is a Lookback of $0.7s$, a horizon of $0.2$ and a RBF bandwidth of $0.5$.}
      \label{tab:diabetes-hyperparameters}
      \begin{tabular}{llrrr}
          \toprule
          Study & Setting & \multicolumn{3}{c}{RMSE} \\
          \cmidrule(lr){3-5}
          & & $a_y$ & $\gamma_{xy}$ & Overall \\
          \midrule
          Lookback & 0.2 & 0.0615 & 0.0126 & 0.0444 \\
                   & 0.5 & 0.0113 & 0.0117 & \textbf{0.0115} \\
                   & 0.7 & 0.0371 & 0.0154 & 0.0284 \\
                   & 0.9 & 0.0477 & 0.0217 & 0.0370 \\
          \midrule
          Horizon  & 0.1 & 0.0239 & 0.0094 & 0.0182 \\
                   & 0.2 & 0.0398 & 0.0105 & 0.0291 \\
                   & 0.3 & 0.0236 & 0.0081 & 0.0177 \\
          \midrule
          RBF bandwidth
                   & 0.3 & 0.0429 & 0.0271 & 0.0359 \\
                   & 0.5 & 0.0398 & 0.0105 & 0.0291 \\
                   & 0.8 & 0.0211 & 0.0065 & 0.0157 \\
          \midrule
          MLE baseline & --- & 0.1188 & 0.0763 & 0.0998 \\
          \bottomrule
      \end{tabular}
  \end{table}

We first center the SDE system, then apply our nonlinear estimator using a basis function expansion and the signature kernel to estimate the target parameters (\cref{eq:nonlin_closed_form}), and then transform the estimated parameters back to the original coordinate system.

We compute the final estimation error over the parameters after transforming back to the original coordinates.

\xhdr{Result and ablation.} Furthermore, we run a small ablation over the parameters of the proposed nonlinear estimation method, specifically, the lookback, the residual integration horizon and the RBF bandwidth, as shown in~\cref{tab:diabetes-hyperparameters}. Overall, the results clearly show that the proposed IV-SDE method can estimate the causal effect with a lower error compared to the biased MLE estimate.
Furthermore, the results remain stable across all settings, demonstrating the method's robustness across reasonable parameter ranges.

Although this SDE system does not fulfill the linear growth assumptions of our theoretical setup defined in~\cref{sec:preliminaries}, the simulated trajectories remain numerically stable under the chosen parameter settings, and our proposed estimator accurately retrieves the objective function. This shows that the estimator may also be applicable beyond the stated regularity conditions.

\xhdr{Alternative examples.} We note that further examples for confounded dynamic settings fulfilling the main IV assumptions can be found e.g. a thermodynamic model of two bodies, where the thermal conductivity between them is the estimand, a known heat source drives one body, and the ambient temperature acts as a common confounder. Another motivating example concerns the interaction between two brain regions, where the stimulation of one region by an e.g. optogenic external impulse acts as the instrument process, while signal from a third region acts as a common confounder.

\end{document}